\documentclass[journal]{IEEEtran}
\usepackage{cite}

\ifCLASSINFOpdf
  \usepackage[pdftex]{graphicx}
    \graphicspath{{./pdf/}{./jpeg/}{./png/}{./eps/}}
      \DeclareGraphicsExtensions{.pdf,.jpeg,.png}
\else
        \usepackage[dvips]{graphicx}
    \graphicspath{{./eps/}}
      \DeclareGraphicsExtensions{.eps}
\fi

\usepackage{tikz}
\usetikzlibrary{shapes,arrows,positioning,calc}
\usepackage{pgfplotstable}
\usepackage{pgfplots}

\usepackage[cmex10]{amsmath}
\usepackage{amsfonts}
\usepackage{amssymb}
\usepackage{mathabx}
\usepackage{mathtools}
\usepackage{algorithmic}
\usepackage{array}
\usepackage{multirow}

\ifCLASSOPTIONcompsoc
  \usepackage[caption=false,font=normalsize,labelfont=sf,textfont=sf]{subfig}
\else
  \usepackage[caption=false,font=footnotesize]{subfig}
\fi

\usepackage{fixltx2e}
\usepackage{url}
\usepackage{hyperref}
\usepackage[absolute]{textpos}
\TPMargin{5pt}

\newcommand{\copyrightstatement}{
    \begin{textblock}{0.96}(0.02,0.95)             \noindent
         \footnotesize
         \copyright  2014 IEEE. Personal use of this material is permitted. Permission from IEEE must be obtained for all other uses, in any current or future media, including reprinting/republishing this material for advertising or promotional purposes, creating new collective works, for resale or redistribution to servers or lists, or reuse of any copyrighted component of this work in other works.
    \end{textblock}
}

\newcommand{\initialstatement}{
    \begin{textblock}{0.85}(0.07,0.01)             \noindent
         \footnotesize
         \centering
         This is the author's version of an article that has been published in this journal. Changes were made to this version by the publisher prior to publication. The final version of record is available at \url{http://dx.doi.org/10.1109/JBHI.2014.2361659}
    \end{textblock}
}

\begin{document}
\initialstatement
\copyrightstatement
\title{A method for context-based adaptive QRS clustering in real-time}

\author{Daniel~Castro*,~Paulo~F\'elix, and Jes\'us~Presedo\thanks{\emph{Asterisk indicates corresponding author.}}
\thanks{*D.~Castro, P.~F\'elix and J.~Presedo are with the CITIUS (Centro de Investigaci\'on en Tecnolox\'ias da Informaci\'on), University of Santiago de Compostela, Santiago de Compostela, 15782 Spain (e-mail: daniel.castro@usc.es; paulo.felix@usc.es; jesus.presedo@usc.es).}\thanks{This work was supported by the Spanish Ministry of Science and Innovation (MICINN) under grant TIN2009-14372-C03-03.}}

\maketitle

\begin{abstract}
Continuous follow-up of heart condition through long-term electrocardiogram monitoring is an invaluable tool for diagnosing some cardiac arrhythmias. In such context, providing tools for fast locating alterations of normal conduction patterns is mandatory and still remains an open issue. This work presents a real-time method for adaptive clustering QRS complexes from multilead ECG signals that provides the set of QRS morphologies that appear during an ECG recording. The method processes the QRS complexes sequentially, grouping them into a dynamic set of clusters based on the information content of the temporal context. The clusters are represented by templates which evolve over time and adapt to the QRS morphology changes.  Rules to create, merge and remove clusters are defined along with techniques for noise detection in order to avoid their proliferation. To cope with beat misalignment, Derivative Dynamic Time Warping is used. The proposed method has been validated against the MIT-BIH Arrhythmia Database and the AHA ECG Database showing a global purity of 98.56\% and 99.56\%, respectively. Results show that our proposal not only provides better results than previous offline solutions but also fulfills real-time requirements.
\end{abstract}


\IEEEpeerreviewmaketitle

\section{Introduction}

\IEEEPARstart{N}{owadays} the surface electrocardiogram (ECG) is recognized as an invaluable tool for monitoring heart condition, since its analysis provides decisive information that can reveal critical deviations from normal cardiac behavior. Recent developments in mobile sensors and mobile computing have enabled new scenarios for continuous ECG monitoring as an inexpensive tool for the early detection of some cardiac events\cite{Sutton2013}, especially in those cases where symptoms appear intermittently. 

As the monitoring period increases, the interpretation task becomes more time consuming and decision-support tools are needed to help cardiologists to reduce the time spent on it. If a continuous follow-up is required, these tools become imperative. Their main aim is to provide the cardiologists with a summary of all the acquired signals, enhanced with a fast locating of those anomalies detected.

Cardiac arrhythmias are the most relevant among the ECG findings. There are two main sources of arrhythmias: an automatism disorder, that is, a set of alterations in the beat activation point due to changes in its location or activation frequency; or a conduction disorder, that is, an abnormal propagation of the beat wavefront through the cardiac tissue. They both have an effect on the ECG, affecting the beat morphology and/or beat rhythm. In order to support their identification, a method for separating the beats by their activation point and conduction pattern should be provided.

\begin{figure*}[!t]
 \centering
 \includegraphics{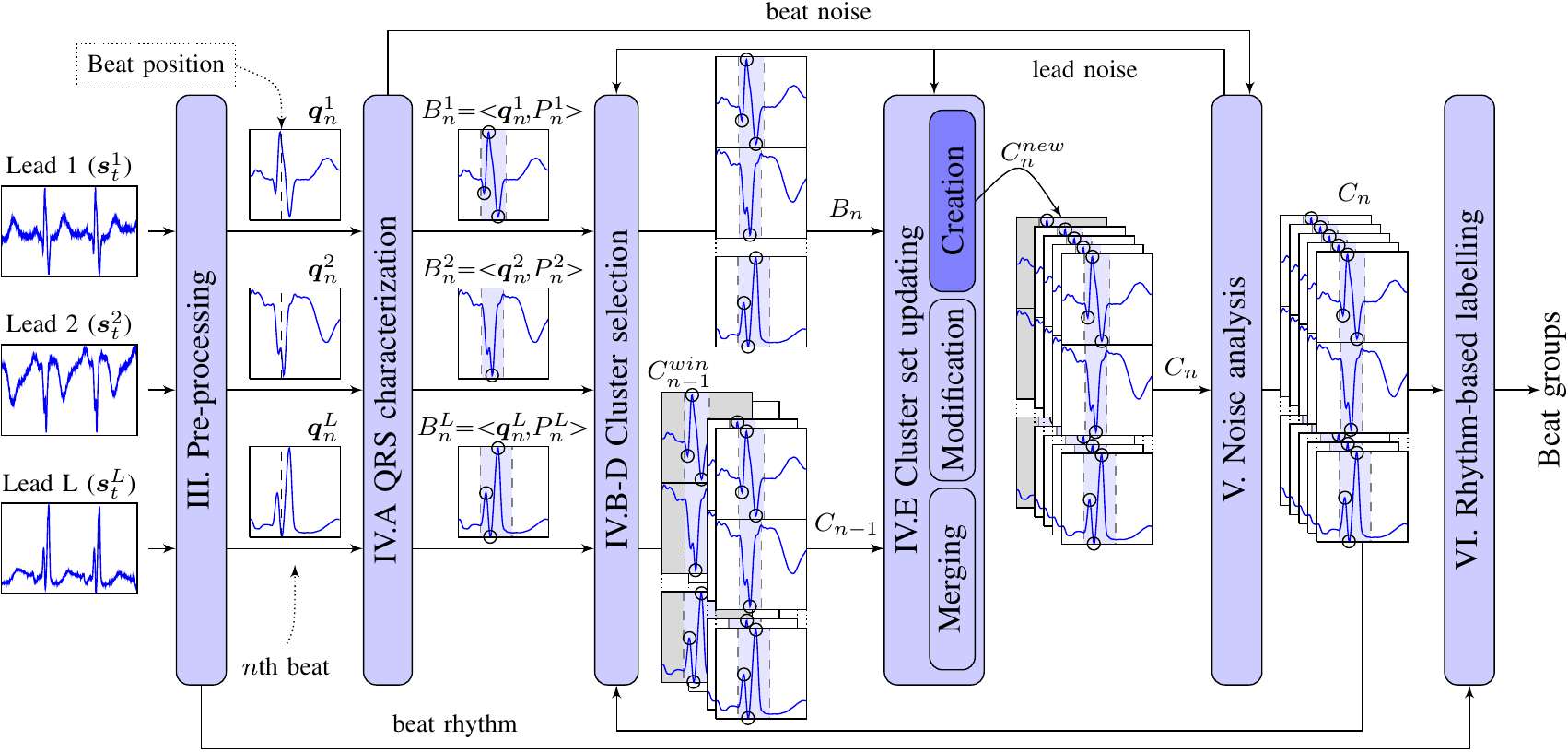}
 \caption{Flow-chart of the method proposed for QRS clustering. The different stages involved in beat processing are shown and the creation of a new cluster is exemplified. Each block refers to the corresponding section.}
\label{fig_scheme}
\end{figure*}

Beat classification arises as the task of assigning each beat in an ECG a label identifying its physiological nature. Machine learning techniques have been applied to this task by estimating the underlying mechanisms that produce the data of a training set. The main drawback of this approach is its strong dependence on the pattern diversity present in the training set. Thus, inter-patient differences show that it cannot be assumed that a classifier trained on data of a large set of patients will yield valid results on a new patient \cite{DeChazal2004,Llamedo2011,deLannoy2012}, and intra-patient differences show that this cannot be assumed even for the same patient throughout time. In addition, class labels only provide gross information about the origin of the beats in the cardiac tissue, loosing all the information about their conduction pathways. This approach does not distinguish the multiple morphological families present in a given class, as occurs in multifocal arrhythmias.

In contrast, beat clustering aims at dividing the ECG recording in a set of beat clusters, each one of them preserving some similarity properties. Previous proposals have focused on an offline approach, from a priori maximum number of clusters \cite{Lagerholm2000,CuestaFrau2003,CuestaFrau2007,RodrguezSotelo2009} and they imply processing the ECG signal once the acquisition has been completed. This approach has given good noise robustness, but as a side effect a single morphology is usually replicated in several clusters and rare beat morphologies can be missed. It also omits the dynamic aspect of ECG and, in particular, ignores the temporal evolution of morphologies. Furthermore, the detection of critical events can be deferred too long to provide timely attention. For all these reasons, a dynamic online approach must be considered.

In this paper, we present a real-time method for adaptive beat clustering, with a potential application not only as a previous step for classification \cite{Krasteva2007}, but also as a summary about those beat morphologies present in a certain period, their temporal evolution and variability, or even to detect the presence of alternating morphologies. The proposed method emulates the experts behavior in exploiting the temporal context for assigning each new beat to the most appropriate cluster. To this end, clusters are continuously adapting to the temporal evolution of beat morphologies, and they can be dynamically created, merged or modified, resulting in a variable number of clusters.

Beat clustering requires extracting from the ECG a set of representative measurements for every beat. Bibliography shows a variety of proposals for beat representation that can be grouped into four categories: {\em morphological features} where the signal amplitude is directly used \cite{Krasteva2007,deLannoy2012,DeChazal2004,CuestaFrau2007}; {\em segmentation features} like area, amplitude or interval duration from beat waves \cite{deLannoy2012,CuestaFrau2007,DeChazal2004,Christov2006} or amplitude and angle values from vectorcardiogram \cite{Llamedo2011}; {\em statistical features} derived using high order cumulants\cite{Osowski2001,Osowski2004,deLannoy2012} and finally, {\em transformed space features}, defined in an alternative space using different transforms like Karhunen\textendash Loewe transform \cite{Hu1997}, Hermite basis functions \cite{Lagerholm2000,Osowski2004,deLannoy2012}, discrete Fourier transform \cite{Dokur2001,Krasteva2007} or wavelet transform \cite{Dokur2001,Llamedo2011}. All the previous works complete their feature sets with rhythm information to complement their description capabilities.

The present paper proposes a new approach to represent a beat by reducing the QRS complex to a set of relevant points and support regions. This representation has some nice properties for beat clustering: it is stable against the usual variability and the presence of noise in the ECG, and it explicitly represents the temporal location of some QRS features. 

The proposed method processes a real-time multilead ECG signal through a set of data-driven stages as shown in Fig.~\ref{fig_scheme}. In order to obtain comparable results, signals from two standard ECG signal databases are used as data source (section \ref{DatabaseSection}). The pre-processing stage comprises real-time beat detection and baseline filtering (section \ref{Preprocessing}). Then, a fixed-length signal segment is selected for extracting and characterizing the QRS complex (subsection \ref{Clustering}.A). QRS complexes are compared to the current set of clusters following a context-based criteria to obtain the best matching cluster (subsections \ref{Clustering}.B\textendash D). Afterwards, the current cluster set is updated in one of three ways: creating a new cluster, modifying the most similar one or merging two or more clusters (subsection \ref{Clustering}.E). The next stage performs a noise analysis for each lead in order to detect noisy intervals and avoid the processing of noisy beats or discard the clusters created from them (section \ref{Noise}). Finally, the beats are classified by their rhythm type and a set of groups with common morphology and rhythm is obtained (section \ref{Rhythm}). 

The ECG databases have been processed and their beat class labels have been used to validate the purity of the final cluster and group sets (section \ref{Results}). 
These results are discussed in section \ref{Discussion} along with the conclusions of the work.

\section{ECG Signal Databases}\label{DatabaseSection}

The ECG databases recommended by the ANSI/AAMI EC57 \cite{EC57-2008} standard for reporting the performance of arrhythmia detectors were used for validation purposes: the MIT-BIH Arrhythmia Database and the AHA ECG database. 

The MIT-BIH Arrhythmia Database \cite{Goldberger2000}\cite{Moody2001} can be referred to as the golden standard for beat clustering and classification tasks and it is the reference database for almost all the literature in this field. This database is composed of 48 recordings of ambulatory ECG, obtained from 47 different patients which comprise a very complete set of examples of common and rare arrhythmias. Each record has a duration of 30 min, and includes two channels with the same leads in almost all of them: a modified-lead II (MLII) in the first one and lead V1 in the second one. MLII was replaced by lead V5 in three records and V1 was replaced by MLII, V2, V4 or V5 on eight records. The signals were digitized at $f_s\!=\!360$Hz and bandpass filtered with cutoff frequencies at 0.1 and 100Hz. All beats present in the database were annotated by at least two expert cardiologists, and assigned a class label using a 16 label set.

The AHA ECG database was compiled by the American Heart Association and it is composed of 155 recordings of ambulatory ECG digitized at $f_s\!=\!250$Hz containing the most relevant types of ventricular arrhythmias. Each record is three hours long with two channels but only the last 30 minutes have been manually annotated by experts.

\section{Pre-processing}\label{Preprocessing}

The major drawback of processing long-term ECG signals is the presence of a high level of noise with multiple manifestations ---baseline wandering, power line interference or electromyographic activity--- so an initial filtering stage needs to be performed. The efforts have been focused on filtering the baseline wandering, as this is the most relevant source affecting the reliability of the clustering algorithm because of the distortion it can cause on the QRS morphology.

\subsection{Baseline filtering}

The baseline filtering is performed through the estimation of the baseline wandering and its posterior elimination from the original ECG. To achieve this goal, each signal is processed by two sequentially connected median filters of 200ms and 600ms, respectively, as described in \cite{DeChazal2004}. A global delay of 400ms is added by this process, independently of the sampling rate.

\subsection{Beat detection}
In order to carry out an evaluation of the clustering method separately from beat detector and provide a fair comparison framework for future algorithms, we used the beat position provided by the ECG databases as fiducial points. 

A beat detector is required for real scenarios, affecting the quality of results. However, it virtually does not increment the global computational complexity as QRS detection represents only a small fraction of it. The global delay would not be affected either, since the beat detection can be performed concurrently with baseline filtering with a shorter delay.

\section{Clustering}\label{Clustering}

In this paper the following notation is used. Bold face variables (e.g., $\boldsymbol{x}$) to represent vectors and sequences, lower case alphabets with subscripts to represent their components (e.g., $x_i$) or superscripts when a temporal index is used as a subscript (e.g., $x_n^i$), upper case alphabets (e.g., $X$) to represent sets, and calligraphic letters to represent functions (e.g., $\mathcal{F}$).

The aim of the clustering method is to group in real-time those beats which share the same activation area and propagation pattern in the cardiac tissue. The propagation of the electrical impulse through the heart is reflected in the ECG as a sequence of waves corresponding to the activity in the atria and ventricles. P wave and QRS complex are of particular interest since they show the atria and ventricular depolarization, respectively, thereby providing a fingerprint of the conduction pathway. Since the noise level of ambulatory signals makes the detection of the P wave a very difficult task, we focus on the QRS morphology to characterize the beat. As a consequence, beats with different atrial activation points or even with nodal activation points can share the same QRS morphology, and will be assigned to the same cluster. In the absence of a P wave analysis, rhythm information can be useful in some cases to detect this situation, as explained in section \ref{Rhythm}.

The proposed method is intended to provide a real-time resume of the ECG signal as a dynamic set of clusters. Whenever a new beat is detected, the set of clusters is updated. Let $C$ denote a time series $C\!=\!\{C_1,C_2,...,C_n,...\}$ where $C_n\!=\!\{C_n^c\!\mid\!1\!\le\!c\!\le\!N_n\}$ is the set of clusters that represents those QRS morphologies which appeared from the beginning of the recording up to the beat $n$, $C_n^c$ is the set of beats assigned to cluster $c$ and $N_n$ the number of clusters, both at beat $n$.   

The clustering follows a data-driven flow triggered by the beat detector as shown in Fig. \ref{fig_scheme}. Each new beat is compared to the set of clusters $C_n$ using concordance and dissimilarity measures in order to find the best matching and giving preference to those in its temporal context. A detailed explanation of this process follows in the remaining subsections: first, a beat characterization and a template-based representation of the clusters; next, the alignment technique and the measures used to compare beats and clusters; afterwards, the criteria for cluster selection and, finally, the process of updating the set of clusters.

\subsection{QRS characterization}\label{characterization}

\begin{figure}[!t]
 \centering
 \includegraphics[width=3.49in]{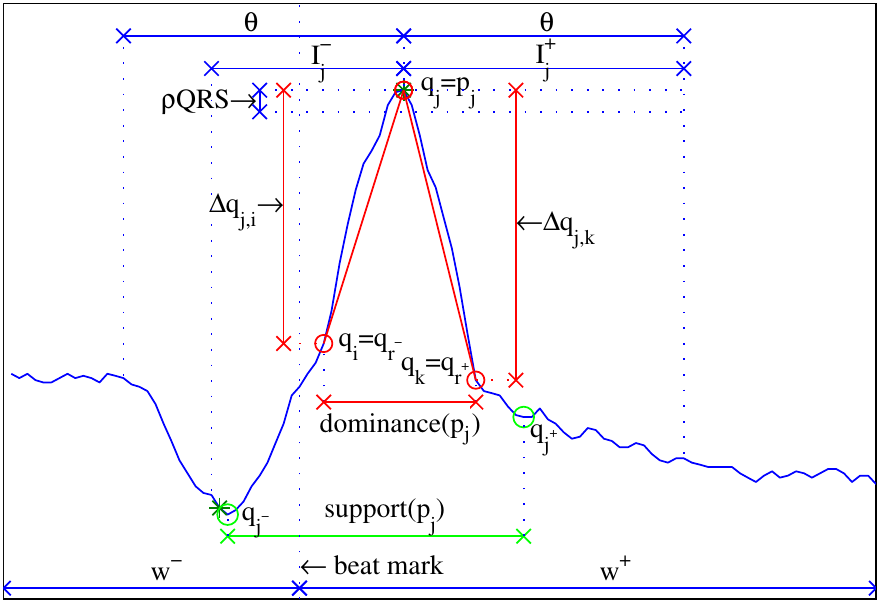}
 \caption{Example of a subsequence $\boldsymbol{q}^l_n$ for lead $l\!=\!2$, of a beat at sample 4102, $n\!=\!13$, from record 108 of the MIT-BIH database. The relevant points detected by the algorithm are marked with *. The parameters involved in the relevant point detection are also shown.}
\label{fig_beatDP}
\end{figure}

We adopt a strategy inspired by dominant point detection \cite{Wu2003} to characterize the QRS morphology through its constituent waves. Let $S\!=\!\{\boldsymbol{s}_t\!\mid\!\boldsymbol{s}_t\!\in\!\mathbb{R}^L \wedge t\!\in\!\mathbb{N}\}$ denote a time series which represents the multilead ECG signal, where $L$ is the number of leads and $\boldsymbol{s}_t\!=\!(s_{t}^1,...,s_{t}^L)$ is the vector of samples at time $t$ for all leads. When the $n$th beat is detected with fiducial mark at time $t$, a fixed-length subsequence of $w\!=\!w^-\!\!+w^+$ samples ($w^-$\! before and $w^+\!\!-\!1$ after $t$) is selected from each lead $l$ to represent its QRS complex (see Fig. \ref{fig_beatDP}):
\begin{equation}
\boldsymbol{q}_n^l=\{q_1^l,\dots,q_j^l,\dots,q_w^l\} 
\end{equation}
where $q_j^l\!=\!s_{t-w^-\!-1+j}^l$. We set $w^-\!\!=\!\lceil0.1\!\times\!f_s\rceil$ and $w^+\!\!=\!\lceil0.2\!\times\!f_s\rceil$, which are wide enough to capture the longest QRS of abnormal beats \cite{CSE1985} (V, F and f beat types, typically).

Most of this section describes operations over one lead, so for the sake of notation simplicity the $l$ superscript will be obviated unless multiple leads are involved.

We define the \textbf{curvature} $\mathcal{K}$ at $q_j\!\in\!\boldsymbol{q}_n$ as:
\begin{equation}
\mathcal{K}(q_j,\boldsymbol{q}_n)= \max_{i \in I_j^-\!, k \in I_j^+}\!\! \cos 
\widehat{q_iq_jq_k},
\end{equation} 
where $2\!\le\!j\!\le\!w\!-\!1$. The terms $I_j^-$ and $I_j^+$ denote the intervals used for calculating the curvature, and they are given by:
\begin{align}
I_j^-\!=\!\{ i \mid\, &(j\!-\!\theta)\!\le\!i\!<\!j\ \wedge \nonumber \\
& \forall a\!\in\!(i,j),(\max_{b \in (a,j)} \Delta q_{j,b}\!-\!\Delta q_{j,a})\!<\!\rho_{min}\}\\
I_j^+\!=\!\{ k \mid\, &j\!<\!k\!\le\!(j\!+\!\theta)\ \wedge \nonumber \\
&\forall a\!\in\!(j,k),(\max_{b \in (j,a)} \Delta q_{j,b}\!-\!\Delta q_{j,a})\!<\!\rho_{min}\}
\end{align}
where $\Delta q_{j,x}\!=\!|q_j\!-\!q_x|$. The term $\theta$ is the maximum physiologically meaningful width of a QRS wave between its peak location and its left or right end, so that it is the upper limit of $I_j^-$ and $I_j^+$. The term $\rho_{min}$ is the minimum height for a signal deflection to be considered physiologically relevant and, therefore, to be excluded from the calculation of curvature. 

We define the \textbf{dominance region} of a sample $q_j$ as:
\begin{equation}
dominance(\!q_j\!)\!=\![r^-\!,r^+], 
\end{equation}
where $r^-\!=\!\min(\operatorname{arg\,max}_{i \in I_j^-} \cos \widehat{q_iq_jq_k})$ for any fixed $k\!\in\!I_j^+$ and $r^+\!=\!\max(\operatorname{arg\,max}_{k \in
I_j^+} \cos \widehat{q_iq_jq_k})$ for any fixed $i\!\in\!I_j^-$.

We define the set of \textbf{dominant points} of $\boldsymbol{q}_n$ as:
\begin{equation}
 D_n\!=\!\{p_j\!\mid\!p_j\!=\!q_j \wedge  j\!=\!\!\!\!\operatorname*{arg\,max}_{a \in dominance(q_j)}\!\!\!\!\mathcal{K} (q_a,\boldsymbol{q}_n) \wedge \Delta q_j\!>\!\rho_{min}\},
\end{equation} 
where $\Delta q_j\!=\! \min(\Delta q_{j,r^-},\Delta q_{j,r^+})$.

We define the set of \textbf{relevant points} of $\boldsymbol{q}_n$ as:
\begin{equation}
R_n\!=\!\{p_j\!\mid\!p_j\!\in\!D_n \wedge \Delta q_j\!>\!\rho_{Q\!R\!S}\}
\end{equation}
where $\rho_{Q\!R\!S}$ is the minimum height for a signal deflection to be considered a relevant QRS wave, with $\rho_{Q\!R\!S}\!>\!\rho_{min}$. If $R_n\!=\!\emptyset$, then it is redefined as: $R_n\!=\!\{\operatorname*{arg\,max}_{p_j \in D_n} \Delta q_j \}$.

The limits of $dominance(q_j)$ can be located at any point in the edge of a wave. Since we are interested in capturing its full extent, the \textbf{support region} of a relevant point $p_j\!\in\!R_n$ is defined as:
\begin{equation}
support(p_j)\!=\![j^-,j^+] 
\end{equation}
where $j^-\!\!\le\!r^-$ and $j^+\!\!\ge\!r^+$ are the sample numbers nearest to $r^-$ \!and $r^+$ where slope sign changes:
\begin{align}
j^-\!\!=&\!\min_{i\in I_j^-}(\{i\!\mid\!i\!\le\!r^-\!\wedge \forall a\!\in\![i,r^-], \Delta q_{j,a}\!>\!\Delta q_{j,a+1}\}) \\
j^+\!\!=&\!\max_{k\in I_j^+}(\{k\!\mid\!k\!\ge\!r^+\!\wedge \forall a\!\in\![r^+,k], \Delta q_{j,a}\!<\!\Delta q_{j,a+1}\}).
\end{align}
In consequence, adjacent support regions can now overlap.

A relevant point $p_j$ is said to be in a concave wave if $q_j\!\!>\!q_{j^-}$ and $q_j\!\!>\!q_{j^+}$. Otherwise, it is said to be in a convex wave. The wave height is defined as $\Delta p_j\!=\! \min(\Delta q_{j,j^-},\Delta q_{j,j^+})$.

Finally, the $n$th beat is represented by the QRS signal segment and the set of its relevant points and support regions $P_n\!=\!\{(p_j,support(p_j))\!\mid\!p_j\!\in\!R_n\}$ and denoted by: 
\begin{equation}
B_n\!=<\boldsymbol{q}_n,P_n\!>.
\end{equation}
 
There is not a consensus in the literature about the limits for width and height of QRS complex or individual QRS waves. The AAMI standard \cite{EC13-2007} recommends a minimum amplitude of $50 \mu V$ and duration of $10$ms for a QRS wave to be detected, and $150 \mu V$ for peak-to-peak QRS amplitude, with a minimum duration of $70 ms$. On the other hand, the AHA \cite{Kligfield2007} and CSE \cite{CSE1985} report lower amplitude for QRS waves (down to $20 \mu V$ and $10$ms), based on measures over averaged beats with increased Signal-to-Noise ratio. These limits were not established for physiological reasons, but for signal noise level or instrumentation limitations. Nothing is stated about maximum QRS width beyond a reference to case-based duration values (e.g. the CSE study \cite{CSE1985} reports a maximum QRS width of $210$ms). In our case, due to the Signal-to-Noise ratio present in the ambulatory signals, the value of $\rho_{Q\!R\!S}$ is set to 150$\mu$V in order to avoid the detection of small waves 
caused by noise and the value of $\theta$ is set to $100$ms to accept QRS waves with a maximum width of $200$ms. The value of $\rho_{min}$ is set to $50\mu V$ following the AAMI standard \cite{EC13-2007} and will be useful to detect noise  contaminated QRS complexes.

In order to perform the comparison between a new beat $B_n$ and the current set of clusters $C_{n-1}$, each cluster $C_{n-1}^c$ is represented by a template:
\begin{equation}
T_{n-1}^c\!=<\!\boldsymbol{q}_{n-1}^c,P_{n-1}^c\!>,
\end{equation}
where $\boldsymbol{q}_{n-1}^c\!=\{q_1^c,\dots,q_w^c\}$ is derived from the QRS of the beats assigned to $C_{n-1}^c$ as will be explained in section \ref{ClusterUpdate}.

\subsection{QRS temporal alignment}

In order to compare a beat $B_n$ with a cluster template $T_{n-1}^{c}$, a temporal alignment of $\boldsymbol{q}_n$ and ${\boldsymbol{q}}_{n-1}^c$ is performed using Dynamic Time Warping (DTW) \cite{Sakoe1978}. The aim of this alignment process is twofold. First, to correct any temporal misalignment due to a misplaced fiducial mark. And second, to reduce the contribution of the height and width variability of the QRS waves to the dissimilarity measure.

DTW was previously used for this purpose in \cite{CuestaFrau2003}, providing a relation $\boldsymbol{m}\!=\!(m_1,...,m_K)$ between $\boldsymbol{q}_n$ and ${\boldsymbol{q}}_{n-1}^c$\! called warping path, with $m_k\!\!=\!\!(x_k,y_k)\!\in\![1,\!w]\!\times\![1,\!w], k\!\in\![1,\!K]$ and $K\!\!\ge\!\!w$. Each $m_k$ represents the alignment of the index $x_k$ of ${\boldsymbol{q}}_n\!$ with the index ${y_k}$ of ${\boldsymbol{q}}_{n-1}^c\!$ under three conditions: first, $m_1\!\!=\!\!(1,\!1)$ and $m_K\!\!=\!\!(w\!,w)$; second, $x_i\!\leq\!x_k$ and $y_i\!\leq\!y_k$ $\forall i\!\!<\!\!k$; and third, $m_{k+1}\!-\!m_k\!\in\!\{(1,\!1),(1,\!0),(0,\!1)\}$. These conditions preserve the time-ordering of points and prevent some point being missed in the alignment.

Let $\mathcal{G}$ denote the cost function associated to a warping path defined by $\mathcal{G}(\boldsymbol{m})\!=\!\sum_{k=1}^K \mathcal{G}_l(x_k,y_k)$, where $\mathcal{G}_l(x,y)\!=\!|q_x\!-q_y^c|$ is the local cost function associated with each element of $\boldsymbol{m}$. The optimal warping path is the one that minimizes $\mathcal{G}$.

DTW aligns the original signal samples, so any component $k$ from $\boldsymbol{m}$ where $m_{k+1}\!-\!m_k\!\in\!\{(1,\!0),(0,\!1)\}$ introduces a horizontal segment into the aligned signals. Since this can lead to unacceptable distortion of the QRS, we adopt the Derivative Dynamic Time Warping (DDTW) \cite{Keogh2001} which uses the estimation of the derivative instead of the signal itself. The derivative is approximated by the first difference: $\dot{\boldsymbol{q}}_n\!=\!(\dot{q}_1,...,\dot{q}_{w-1})$ and $\dot{{\boldsymbol{q}}}_{n-1}^c\!=\!(\dot{q}_1^c,...,\dot{q}_{w-1}^c)$ with $\dot{q}_x\!=\!q_{x+1}\!-\!q_{x}$ and $\dot{q}_y^c\!=\!q_{y+1}^c\!-\!q_{y}^c$.

\begin{figure}[t!]
 \centering
 \includegraphics[width=3.4in]{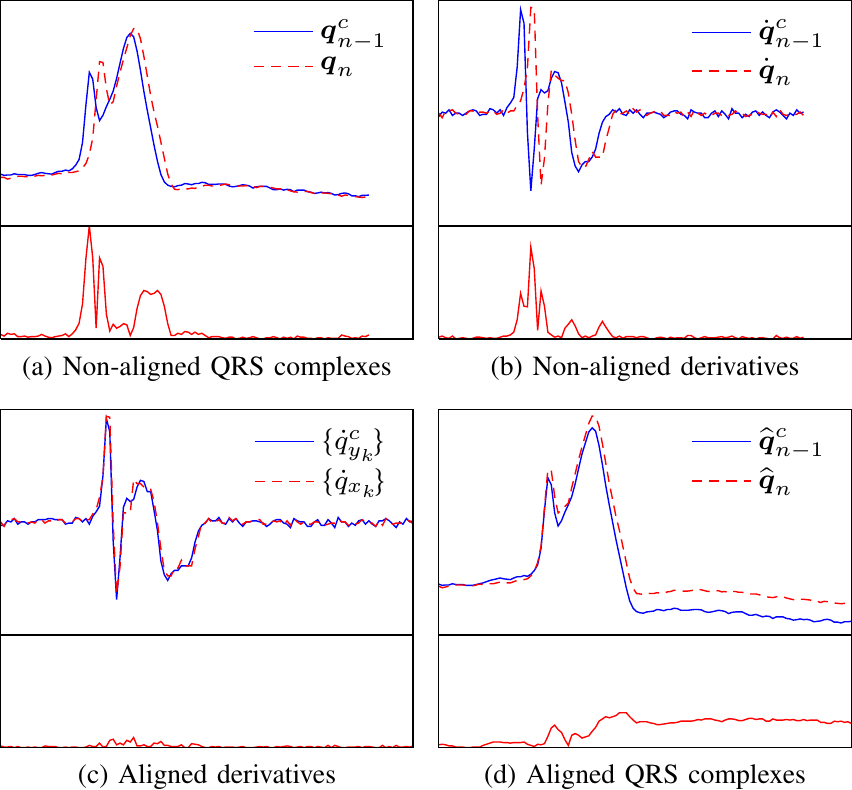}
 \caption{Example of the DDTW alignment process. For each subfigure, the upper solid and dashed lines represent the beat and template data, respectively; bottom solid lines represent their absolute difference. (a) Shows $\boldsymbol{q}_n$ and $\boldsymbol{q}_{n-1}^c$ subsequences; (b) $\dot{\boldsymbol{q}}_n$ and $\dot{\boldsymbol{q}}_{n-1}^c$ derivative approximations; (c) optimally-aligned derivatives of length $K\!\!\ge\!\!w\!-\!1$; (d) subsequences $\widehat{\boldsymbol{q}}_n$ and $\widehat{\boldsymbol{q}}_{n-1}^c$ of length $K\!+\!1$ obtained from the optimally-aligned derivatives. In (b) and (c), the difference is equivalent to the local cost function $\mathcal{G}_l(x,y)\!=\!|\dot{q}_x-\dot{q}_y^c|$. Notice the temporal increment $K\!\!-\!\!w$ in (c) and (d) as a result of the process. The absolute difference may also be increased in (d) outside the support regions, but this is irrelevant.}
\label{fig_sim}
\end{figure}

We imposed some additional conditions on the selected warping path so as to restrict the alterations of aligned signals. A global restriction is set in the search process by defining a warping window $\delta$ (named Sakoe-Chiba band \cite{Sakoe1978}) to limit the temporal distance between aligned samples so $|x_k\!-\!y_k|\!<\!\delta$. We set $\delta\!=\!5$, which corresponds to a distance of 14ms with $f_s\!=\!360$Hz, and is long enough to deal with small misalignments of beat marks. A local restriction is also used to limit the number of times the same sample can be aligned, setting a slope constraint: $m_{k+a}\!-\!m_k\!\notin\!\{(0,a),(a,0)\},\forall a\!>\!\lambda$. We set $\lambda\!=\!2$ to limit the variation in the amplitude of the aligned signals. Both conditions together allows the DDTW to cancel out wave differences up to three times in amplitude and up to $\delta$ samples in width. 

Finally, after the optimal warping path $\boldsymbol{m}$ is found, the aligned signals $\widehat{\boldsymbol{q}}_n$ and $\widehat{{\boldsymbol{q}}}_{n-1}^c$ are obtained with coordinates $\widehat{q}_{k+1}\!=\!\widehat{q}_{k}\!+\!\dot{q}_{x_k}$ and $\widehat{q}_{k+1}^c\!=\!\widehat{q}_{k}^c\!+\!\dot{q}_{y_k}^c $ for $k\!\in\![1,K]$, where
$\widehat{q}_1\!=\!q_1$ and $\widehat{q}_1^c\!=\!q_1^c$. Fig. \ref{fig_sim} shows the result of the alignment of the current QRS complex and a template. 

\subsection{Template matching} \label{dissimilarity}
  
\begin{figure}[!t]
 \centering
 \includegraphics{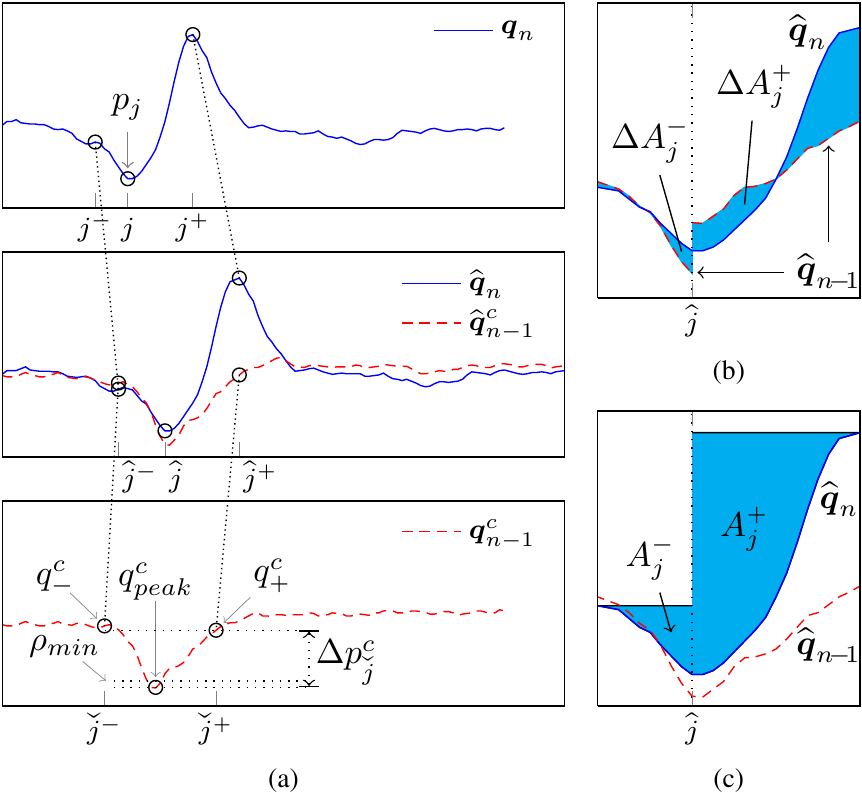}
 \caption{Parameters involved in calculating the concordance and local dissimilarity of a cluster with respect to a beat at a relevant point $p_j$. Solid and dashed lines represent beat and template data, respectively. (a) Shows parameters involved in concordance checking; the support region $[j^-,j^+]$ and the derived intervals $[\widehat{j}^-,\widehat{j}^+]$ and $[\widecheck{j}^-,\widecheck{j}^+]$ are drawn; (b) signal segments within the interval $[\widehat{j}-,\widehat{j}+]$ with a vertical alignment performed in the subintervals $[\widehat{j}^-,\widehat{j}]$ and $[\widehat{j},\widehat{j}^+]$ independently; $\Delta A_j^-$ and $\Delta A_j^+$ areas are shaded; (c) the same signal segments shown in (b) with $A^-_j$ and $A^+_j$ areas shaded.}
\label{fig_dissimilarity}
\end{figure}

In order to assign a beat $B_n$ to an existent cluster $C_{n-1}^c$, we design a similarity calculation that only considers the difference between signals over the support region of each relevant point, thus limiting the comparison to the constituent waves of the QRS. Given a pair $(p_j,[j^-,j^+])\!\in\!P_n$, we are interested in verifying whether $T_{n-1}^c$ contains a similar wave in the same location of the QRS. In order to do so, the interval $[\widecheck{j}^-,\widecheck{j}^+]$ of $\boldsymbol{q}_{n-1}^c$ aligned with the interval $[j^-,j^+]$ of $\boldsymbol{q}_n$ must be obtained (see Fig. \ref{fig_dissimilarity}). To this end, the warping path $\boldsymbol{m}$ is  used to map $j$ and $[j^-\!,j^+\!]$ from $\boldsymbol{q}_n$ into $\widehat{\boldsymbol{q}}_n$ obtaining the equivalent index $\widehat{j}$ for the relevant point, and $[\widehat{j}^-\!,\widehat{j}^+]$ for its support region where: $\widehat{j}\!=\!\max\{k \!\mid\! x_k\!\!=\!j\}$, $\widehat{j}^-\!\!=\!\min\{k 
\!\mid\! x_k\!\!=\!j^-\!\}$ and $\widehat{j}^+\!\!=\!\max\{k \!\mid\! x_k\!\!=\!j^+\!\}$, being $(x_{k},y_{k})\!\in\!\boldsymbol{m}$. Afterwards, since $\widehat{\boldsymbol{q}}_{n-1}^c$ and $\widehat{\boldsymbol{q}}_n$ are already aligned by the application of DDTW, the same interval $[\widehat{j}^-\!,\widehat{j}^+]$ from $\widehat{\boldsymbol{q}}_{n-1}^c$ is selected and mapped into ${\boldsymbol{q}}_{n-1}^c$ using the warping path. The interval $[\widecheck{j}^-,\widecheck{j}^+]$ is given by $\widecheck{j}^-\!\!=\!y_{\widehat{j}^-}$ and $\widecheck{j}^+\!\!=\!y_{\widehat{j}^+}$.

Once the aligned intervals are obtained we proceed to evaluate the concordance of both vectors. The segment ${\boldsymbol{q}}_{n-1}^c$ is said to \textbf{concord with 
${\boldsymbol{q}}_n$ at $p_j$} and denoted by ${\boldsymbol{q}}_{n-1}^c\!\approx_{p_j}\!\!{\boldsymbol{q}}_n$ if ${\boldsymbol{q}}_{n-1}^c$ contains a deflection in $[\widecheck{j}^-\!,\widecheck{j}^+\!]$ with height $\Delta p^c_{\widecheck{j}} \!>\!\rho_{min}$ likely to be considered a significant waveform, where $\Delta 
p^c_{\widecheck{j}}\!=\!\min(|q_{peak}^c\!-q^c_-|,|q_{peak}^c\!-q_+^c|)$ being $peak\!=\!\operatorname{arg\,min}_{i\in[\widecheck{j}^-\!,\widecheck{j}^+\!]} q_i^c$, 
$q_-^c\!=\!\max_{i\in[\widecheck{j}^-\!,peak]} q_i^c$ and $q_+^c\!=\!\max_{i\in[peak,\widecheck{j}^+]} q_i^c$ for $p_j$ in a convex wave (see Fig. \ref{fig_dissimilarity}a). Otherwise, $\min$ and $\max$ functions are interchanged.

We define the \textbf{concordance ratio} of ${\boldsymbol{q}}_{n-1}^c$with respect to ${\boldsymbol{q}}_n$ at $p_j$, denoted by $\mathcal{C}_{p_j}({\boldsymbol{q}}_{n-1}^c,{\boldsymbol{q}}_n)$, as: 
\begin{equation}
\mathcal{C}_{p_j}({\boldsymbol{q}}_{n-1}^c,{\boldsymbol{q}}_n)\!=\!
\frac{\min(\Delta p_j, \Delta p^c_{\widecheck{j}})}{\max(\Delta p_j, \Delta p^c_{\widecheck{j}})}
\end{equation}
if ${\boldsymbol{q}}_{n-1}^c\!\approx_{p_j}\!\!{\boldsymbol{q}}_n$. Otherwise, $\mathcal{C}_{p_j}({\boldsymbol{q}}_{n-1}^c,{\boldsymbol{q}}_n)\!=\!0$.

We define the \textbf{local dissimilarity} of ${\boldsymbol{q}}_{n-1}^c$ with respect to ${\boldsymbol{q}}_n$ at $p_j$, and denoted by $\mathcal{D}_{p_j}({\boldsymbol{q}}_{n-1}^c,{\boldsymbol{q}}_n)$, as a weighted relative area difference between $\widehat{\boldsymbol{q}}_{n-1}^c$ and $\widehat{\boldsymbol{q}}_n$
at the interval $[\widehat{j}^-,\widehat{j}^+]$:
\begin{equation}
\mathcal{D}_{p_j}({\boldsymbol{q}}_{n-1}^c,{\boldsymbol{q}}_n)=
\left (\frac{(\Delta A_j^-)^2}{A_j^-}\!+\!\frac{(\Delta A_j^+)^2}{A_j^+}\right) \times\frac{1}{A_j^-\!+\!A_j^+}
\end{equation}

The terms $\Delta A_j^-$and $\Delta A_j^+$ represent the areas under $\boldsymbol{a}\!=\!|\widehat{\boldsymbol{q}}_n\!-\widehat{\boldsymbol{q}}_{n-1}^c \!|$ over
the intervals $[\widehat{j}^-\!, \widehat j]$ and $[\widehat j,\widehat{j}^+]$, respectively (see Fig. \ref{fig_dissimilarity}b). Computing the area at each side 
of $\widehat j$ independently allows us to minimize the effect of vertical misalignment or amplitude variation on each interval by subtracting their own median. Using the trapezoidal rule:
\begin{align}
\Delta A_j^-\!=\!\sum_{k\in(\widehat{j}^-\!,\widehat j)}\!\!a_k\!+\!\frac{1}{2}(a_{\widehat{j}^-}\!+\!a_{\widehat j})\!-\!(\widehat{j}-\widehat{j}^-)M^-\\
\Delta A_j^+\!=\!\sum_{k\in(\widehat{j},\widehat{j}^+)}\!\!a_k\!+\!\frac{1}{2}(a_{\widehat{j}}\!+\!a_{\widehat{j}^+})\!-\!(\widehat{j}^+\!-\widehat{j})M^+ 
\end{align}
where $M^-\!=\!\operatorname*{median}_{k\in[\widehat{j}^-\!,\widehat j]}\!a_k$ and $M^+\!=\!\operatorname*{median}_{k\in[\widehat j,\widehat{j}^+\!]}\!a_k$.

The terms $A_j^-$ and $A_j^+$ represent the areas below $[\widehat{j}^-\!,\widehat j]$ and $[\widehat j,\widehat{j}^+]$ intervals of $\widehat{\boldsymbol{q}}_n$, respectively (see Fig. \ref{fig_dissimilarity}c). They are estimated using the trapezoidal rule:
\begin{align}
A_j^-\!=\!\left | \sum_{k\in(\widehat{j}^-\!,\widehat{j})}\!\widehat{q}_k + 
\frac{1}{2}(\widehat{q}_{\widehat{j}^-}\!\!+\!\widehat{q}_{\widehat{j}}) - 
(\widehat{j}\!-\!\widehat{j}^-\!)\widehat{q}^- \right |\\
A_j^+\!=\!\left | \sum_{k\in(\widehat{j},\widehat{j}^+\!)}\!\widehat{q}_k + \frac{1}{2}(\widehat{q}_{\widehat{j}}\!+\!\widehat{q}_{\widehat{j}^+}\!) - (\widehat{j}^+\!\!-\!\widehat{j})\widehat{q}^+ \right | 
\end{align}
where $\widehat{q}^-\!=\!\max_{k\in[\widehat{j}^-\!,\widehat j]}\!q_k$ and $\widehat{q}^+\!=\!\max_{k\in[\widehat j,\widehat{j}^+\!]}\!q_k$ for $p_j$ in a convex wave. Otherwise, $\max$ is replaced by $\min$.

We define the \textbf{piecewise similarity} of ${\boldsymbol{q}}_{n-1}^c$\! with respect to ${\boldsymbol{q}}_n$, denoted $\mathcal{P\!S}({\boldsymbol{q}}_{n-1}^c,{\boldsymbol{q}}_n)$, as the sum of two bounded contributions from the set of concordant and non-concordant relevant points:
\begin{align}
 \mathcal{P\!S}(\boldsymbol{q}_{n-1}^c,\boldsymbol{q}_n)=&
\sum_{p_j \in R_n}
\mathcal{C}_{p_{j}}(\boldsymbol{q}_{n-1}^c,\boldsymbol{q}_n)
.\operatorname{sig}(\mathcal {D}_{\!p_{\!j}}(\boldsymbol{q}_{n-1}^c,\boldsymbol{q}_n)) \nonumber \\ 
& {} - \!\!\!\max_{p_{\!j}\mid\boldsymbol{q}_{n-1}^c\!\not\approx_{p_{\!j}}\!\boldsymbol{q}_n} 
\!\!\!\!\mathcal{D}_{\!p_{\!j}}\!(\!\boldsymbol{q}_{n-1}^c,\!\boldsymbol{q}_n\!) 
\end{align}
where the sigmoid function $\operatorname{sig}(x)\!\!=\!\!1\!-\!\alpha x / \!\sqrt{1\!+\!(\alpha x)^2}$ keeps the contribution of each point in the interval $[0,1]$. The parameter $\alpha$ determines the decrease rate of the contribution as the local dissimilarity increases and, as a consequence, the weight of a high dissimilarity value in the final piecewise dissimilarity. We limit the contributions of those points out of a range of admissible local dissimilarity. In order to set such range we consider the effect of amplitude and temporal variability of QRS waves. A temporal misalignment of one signal sample between the QRS waves of $\boldsymbol{\widehat q}_n$ and $\boldsymbol{\widehat q}_{n-1}^{c}$ can lead to local dissimilarities of around 20\%. Then we set an interval of $[0,25\%]$ with a slightly greater upper bound and $\alpha\!=\!4$ so the maximum contribution out of this interval is $0.30$.

The previous measure is asymmetric since it depends on the relevant points and areas of one signal, so we define the \textbf{similarity} between $B_n$ and 
$T^c_{n-1}$ as: 
\begin{equation}
\mathcal{S}(\boldsymbol{q}_n,\boldsymbol{q}_{n-1}^c)\!=\!\mathcal{P\!S}
(\boldsymbol{q}_{n-1}^c, 
\boldsymbol{q}_n)+\mathcal{P\!S}(\boldsymbol{q}_n,\boldsymbol{q}_{n-1}^c)
\end{equation}
thus obtaining a value which captures the concordance, similarity and morphological complexity of both signal segments.

This measure allows us to select the most similar template within a set, but we need a reference scale to evaluate the degree of matching. To this end, we define
the \textbf{normalized piecewise similarity} as:
\begin{equation} 
\widebar{\mathcal{P\!S}}(\boldsymbol{q}_{n-1}^{c},\boldsymbol{q}_{n})\!=\!\mathcal{P\!S}(\boldsymbol{q}_{n-1}^{c},\boldsymbol{q}_{n})/|R_{n}|
\end{equation}
and the \textbf{normalized similarity} as:
\begin{equation}
\bar{\mathcal{S}}(\boldsymbol{q}_n,\boldsymbol{q}_{n-1}^{c}
)\!=\!\mathcal{S} 
(\boldsymbol{q}_n,\boldsymbol{q}_{n-1}^{c})/(|R_n|+|R^{c}_{n-1}|),
\end{equation}
where $R_{n-1}^{c}$ is the set of relevant points of $\boldsymbol{q}_{n-1}^{c}$.

\subsection{Cluster selection}\label{ClusterSelection}

\begin{figure}[!t]
 \centering
 \includegraphics{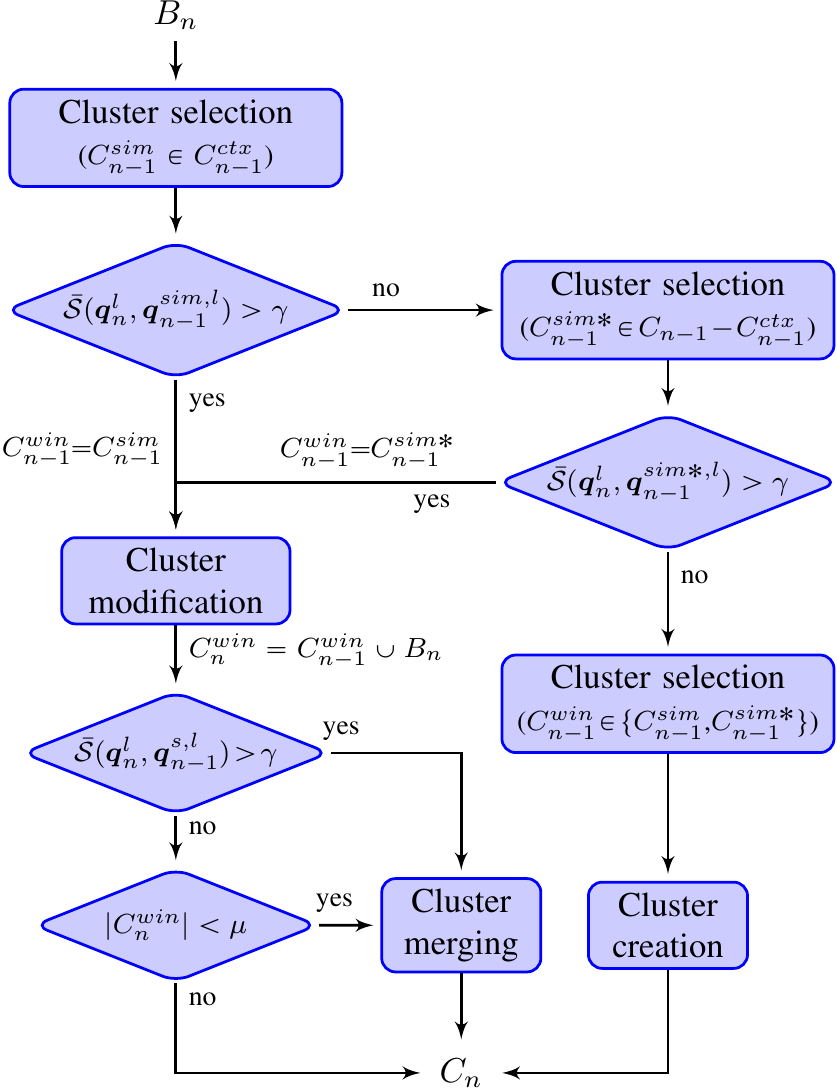}
\caption{Flow-chart of the cluster selection and cluster set updating processes.}
\label{fig_cluster_selection}
\end{figure}

The occurrence of different QRS morphologies in a segment of ECG signal is usually limited by a reduced set of activation points and conduction pathways. Thereby we expect that the majority of QRS complexes in an ECG recording share their morphology with some of the QRS complexes present in a short previous temporal interval. For that reason, the search for the cluster $C_{n-1}^{win}\!\in\!C_{n-1}$ that best matches a beat $B_n$ is first performed in the set of clusters present in its temporal context $C_{n-1}^{ctx}\!\subset\!C_{n-1}$ (see Fig. \ref{fig_cluster_selection}). We define the temporal context as the set of $\tau$ beats previous to $B_n$, $\tau\text{\textendash}ctx^-(B_n)\!=\!\{B_{n-i}\!\mid\!1\!\le\!i\!\le\!\tau\}$ and $C_{n-1}^{ctx}\!=\!\{C_{n-1}^c \!\in\!C_{n-1} \!\mid\! \exists B_i\!\in\!\tau\text{\textendash}ctx^-(B_n)\!\wedge\!B_i\!\in\!C_{n-1}^c \}$. The context length is set to $\tau\!=\!15$ beats, the number of beats displayed in the typical 10s ECG strip used by cardiologists for a heart rate of 80 beats/min. This context is long enough to include every QRS morphology present in multifocal arrhythmias. Throughout this section, the $l$ superscript is used to denote the lead. 

The similarity measure is used to identify the most similar template for each lead as $sim^l\!=\!\operatorname{arg\ max}_c \mathcal{S}(\boldsymbol{q}_n^l,\boldsymbol{q}_{n-1}^{c,l})$ and then the best matching cluster is obtained by majority vote as $sim\!=\!\operatorname{mode}\{sim^l\!\mid\!l\!\in\![1,L]\}$. If multiple clusters are selected, a second vote is performed to obtain a single one using the normalized similarity.

Beat $B_n$ is assigned to $C_{n-1}^{sim}$ if the condition $\bar{\mathcal{S}}(\boldsymbol{q}_n^l,\boldsymbol{q}_{n-1}^{sim\!,l})\!\!>\!\!\gamma$ is fulfilled in all leads. Then $C_{n-1}^{win}\!\!=\!C_{n-1}^{sim}$.  We set a value of $\gamma\!=\!0.30$ which corresponds to the maximum contribution of a point with local dissimilarity outside the admissible interval.

When $B_n$ and $C_{n-1}^{sim}$ are not similar enough, a new comparison is performed within the subset $C_{n-1}\!-C_{n-1}^{ctx}$, obtaining the most similar cluster $C_{n-1}^{sim*}$. If $B_n$ and $C_{n-1}^{sim*}$ are similar enough, the beat is assigned and $C_{n-1}^{win}\!\!=\!C_{n-1}^{sim*}$. Otherwise, the beat is not assigned to any existing cluster and its most similar cluster $C_{n-1}^{win}$ is selected between $C_{n-1}^{sim}$ and $C_{n-1}^{sim*}$ using the same voting criteria previously seen.

\subsection{Cluster set updating} \label{ClusterUpdate}

In order to adaptively respond to the changing behavior of the ECG, clusters must be dynamically created, modified or merged whenever a new beat is detected:

\subsubsection{Cluster creation} If $B_n$ is not assigned to $C_{n-1}^{win}$, a new cluster $C_{n}^{new}$ is created and its template is initialized for each lead using the beat representation: $T_{n}^{new,l}\!=\!B_n^l$. Then the cluster set is updated as $C_{n}\!=\!C_{n-1}\!\cup \{C_{n}^{new}\}$.

\subsubsection{Cluster modification}
If $B_n$ is assigned to $C_{n-1}^{win}$, then the cluster is updated to $C_{n}^{win}\!\!=\!C_{n-1}^{win}\!\cup \{B_n\}$ and the template $T_{n-1}^{win,l}\!\!=<\!\boldsymbol{q}_{n-1}^{win,l},P_{n-1}^{win,l}\!>$ is modified to $T_{n}^{win,l}$ for each lead. To this end, $\boldsymbol{q}_{n}^{win,l}$ is calculated from $\boldsymbol{q}_n^l$ and $\boldsymbol{q}^{win,l}_{n-1}$:
\begin{align}\label{eq_update}
&\dot{q}_{y,n}^{win.l}\!=\!(1-\beta) \dot{q}_{y,n-1}^{win,l} + \beta
\operatorname{mean}(\{\dot{q}_{x,n}^l\!\mid\!(x,y)\!\in\!\boldsymbol{m}\})  \\
&q_{y,n}^{win,l}\!=\!q_{y-1,n}^{win,l}+\dot{q}_{y,n}^{win,l}
\end{align}
where $q_{1,n}^{win,l}\!=\!q_{1,n-1}^{win,l}$. The term $\beta$ is the coefficient of the exponential update. Setting a value for $\beta$ implies a trade-off between
plasticity and stability of the cluster template. We set $\beta = 1/8$ so the last 16 beats assigned to the cluster provide 90\% of the contributions to the current template. This allows the template to be adapted to the evolution of the QRS morphology. Afterwards, the set of relevant points and support regions $P_{n}^{win,l}$ is obtained from $\boldsymbol{q}_{n}^{win,l}$ and assigned to the template: $T_{n}^{win,l}\!=<\!\boldsymbol{q}_{n}^{win,l},P_{n}^{win,l}\!>$. 

\subsubsection{Cluster merging} During the clustering process, different clusters can evolve to represent the same QRS morphology, so they should be merged. This situation is common when QRS complexes suffer from transient distortions in their morphology due to intrinsic variability, which makes them fall below the similarity threshold for their proper clusters. In this case, a new cluster is created which is subjected to an initial transient period during which the template for each lead can evolve getting closer to its most similar cluster. 

In order to detect this situation, we define a relation $closest$ which links each cluster with its most similar one among previous clusters. The relation is set for each new cluster as $C_{n}^{win}\!=\!closest(C_{n}^{new})$. As templates evolve with new assigned QRS complexes, the relation could have to be updated. 

When $B_n$ is assigned to $C_{n-1}^{win}$, the cluster is updated to $C_{n}^{win}$ and its most similar cluster may change. The $closest$ relation is updated when multiple clusters in the same set than $C_{n-1}^{win}$, be it either $C_{n-1}^{ctx}$ or $C_{n-1}\!-\!C_{n-1}^{ctx}$, fulfill the assigment condition $\bar{\mathcal{S}}(\boldsymbol{q}_n^l,\boldsymbol{q}_{n-1}^{c,l})\!>\!\gamma$ in all leads (see subsection \ref{ClusterSelection}). Since $C_n$ has been updated, getting more similar to $B_n$, the best matching cluster $C_{n-1}^s$ for $B_n$ is selected within the remaining subset of clusters, where $s\!=\!\operatorname{arg\ max}_{C_{n-1}^c\ne C_{n-1}^{win}}(\sum_{l=1}^L\mathcal{S}(\boldsymbol{q}^l_{n},\boldsymbol{q}^{c,l}_{n-1})/L)$ and the oldest of both clusters, $C_{n-1}^{win}$ or $C_{n-1}^s$, is set as the new most similar cluster for the other. Then the clusters $C_{n}^{win}$ and $C_{n}^s$ are checked for possibly merging, since their templates have evolved to be similar enough to the last beat (see Fig. \ref{fig_cluster_selection}). The condition for merging those clusters will be analogous to the condition for beat assignment: $\bar{\mathcal{S}}(\boldsymbol{q}_n^{s,l},\boldsymbol{q}_n^{win,l})\!>\!\gamma'$, where $\gamma'\!=\!0.40$, which corresponds to the maximum contribution of a point with local dissimilarity over 20\%. The threshold value is increased with respect to $\gamma$ since both signal segments are promediated templates with increased Signal-to-Noise ratio.
 
Afterwards, the special case of clusters within its transient period is considered. We establish the duration of this transitory state in terms of the number of assigned beats. Hence, if $B_n$ is assigned to $C_{n-1}^{win}$ and $|C_{n}^{win}|\!<\!\mu$, the cluster is checked for merging with its closest one $C_{n}^s\!=\!closest(C_{n}^{win})$ (see Fig. \ref{fig_cluster_selection}). We set $\mu$ as the minimum number of beats assigned to the cluster to confirm it represents an independent morphology and we consider $\mu\!=\!10$ enough for this purpose. 

When two clusters $C^{win}_n$ and $C_n^c$ are merged, the cluster set, cluster template and $closest$ relation are updated accordingly (we suppose that $C_n^c\!\!=\!closest(C_n^{win})$): 
\begin{enumerate}
 \item $C^{c}_{n}$ is updated to $C^{c}_{n}\!=\!C^{c}_{n}\cup C^{win}_{n}$.
 \item $\boldsymbol{q}_{n}^{c,l}$ is calculated by merging $\boldsymbol{q}^{c,l}_{n-1}$ and $\boldsymbol{q}^{win,l}_n$:
\begin{align}
&\dot{q}_{y,n}^{c,l}\!=\!(1\!-\!\beta) \dot{q}_{y,n}^{c,l}\! +\! \beta
\sum(\{\dot{q}^{win,l}_{x,n}\!\mid\!(x,y)\!\in\!\boldsymbol{m}\})  \\
&q_{y,n}^{c,l}\!=\!q_{y-1,n}^{c,l}+\dot{q}_{y,n}^{c,l}
\end{align}
where $q_{1,n}^{c,l}$ remains unmodified.
\item The template is modified to $T_{n}^{c,l}\!=<\!\boldsymbol{q}_{n}^{c,l},P_{n}^{c,l}\!>$, where
$P_{n}^{c,l}$ is the set of relevant points and support regions obtained  of $\boldsymbol{q}_{n}^{c,l}$.
\item The $closest$ relation is updated by removing the $(C_n^c,C_n^{win})$ pair and replacing $C_n^{win}$ by $C_n^{c}$ in all the pairs where the former appears.
\end{enumerate}

After the cluster $C_n^{win}$ gets merged, all the modified pairs of the $closest$ relation are checked for merging. Afterwards, the $C_n^c$ cluster is also checked for merging with its closest one.

\section{Noise-cluster proliferation control}\label{Noise}

The dynamic creation of clusters entails the problem of identifying those QRS complexes contaminated by noise that could cause the proliferation of clusters. When noise appears in an ECG lead, the QRS can show different changes. In some cases, ECG segments with low Signal-to-Noise ratio present a high number of waves, and these  can be detected by an abnormal number of dominant points. In other cases, domain knowledge is needed to discern if changes respond to a new QRS morphology or to a noisy version of a previous one. Some alterations are well known and described in literature, but others can be challenging even for an expert cardiologist, who compares every beat
with those present in its temporal context in order to identify if it is related to a repetitive morphology change, a noisy interval or an isolated noisy beat. We follow a twofold, beat-based and context-based, approach to detect noisy QRS complexes. Noise can be present in one or more leads, and the influence of each lead in the clustering of a given beat will be analyzed in the following. 

A state variable $lead\_noise_n^l$ is defined to denote the existence of a \emph{noisy interval} in lead $l$ containing the $n$th beat. Such noisy interval begins when the first noisy beat is detected in that lead just after a sequence of previous noise-free beats. The beats contained within this interval can be either noisy or noise-free in $l$ and such condition will be represented as an attribute of the beat denoted by $beat\_noise_n^l$. During the noisy interval, the characterization of any new beat $B_n$ can represent not only QRS waves but also noise artifacts, so the cluster selection and assignment rules (subsection \ref{ClusterSelection}) are modified in lead $l$: the dominant points of the beat are ignored, replacing $\mathcal{S}(\boldsymbol{q}^l_n,\boldsymbol{q}_{n-1}^{c,l})$  by $\mathcal{P\!S}(\boldsymbol{q}^l_n,\boldsymbol{q}_{n-1}^{c,l})$ and $\bar{\mathcal{S}}(\boldsymbol{q}^l_n,\boldsymbol{q}_{n-1}^{c,l})$ by $\widebar{\mathcal{P\!S}}(\boldsymbol{q}^l_n,\boldsymbol{q}_{n-1}^{c,l})$. Therefore, the condition for beat assignment is: $\widebar{\mathcal{P\!S}}(\boldsymbol{q}^l_n,\boldsymbol{q}_{n-1}^{win,l})\!>\!\gamma$. 

The ending of a noisy interval in lead $l$ is set just before $\kappa\!=\!3$ contiguous beats are considered noise-free in that lead. A beat $B_n$ is considered \emph{noise-free} if $beat\_noise^l_n\!\!=\!\!false$ and it is assigned either to its winner cluster with $\widebar{\mathcal{S}}(\boldsymbol{q}^l_n,\boldsymbol{q}_{n-1}^{win,l})\!>\!\gamma$ or to a new cluster.

The \emph{beat-based noise detection} is triggered when a new beat $B_n$ is detected. Then the number of waves in the QRS is estimated as $|D^l_n|$ for each lead $l$. If $|D_n^l|\!>\!\eta$, $B_n$ is considered noisy in lead $l$ and $beat\_noise_n^l$ is set to $true$. The term $\eta$ is the maximum number of waves that could be present in a noise\textendash free QRS. We set $\eta\!=\!6$ to admit complexes with Q, R and S waves, mixed with R', S' or spikes as occurs in paced, fusion or bundle branch block beats. Additionally, if $|R_n^l|\!>\!\eta$, the QRS complex is considered distorted with relevant waves caused by noise. If this happens in some but not all leads, such leads are ignored for cluster assignment and updating, but if it happens in all leads, the beat is considered failed and it is assigned to its most similar cluster. 

The \emph{context-based noise detection} is activated when a new cluster is created for $B_n$. Then two possible explanations are explored. A first explanation represents a hypothesis of noise, and every noisy beat will be assigned to its most similar cluster. A second explanation simply represents a change in morphology, so creation of new clusters is allowed. The evaluation of these hypothesis is performed over a temporal context of $\tau$ beats defined as $\tau\text{\textendash}ctx^+(B_n)\!=\!\{B_{n+i}\!\mid\!0\!\le\!i\!<\!\tau\}$ which is long enough to check the evolution of cluster diversity. 
 
A hypothesis of noise, denoted by $hyp\_noise^l_{n+i}$, is set over each lead for every beat $B_{n+i}\!\in\!\tau\text{\textendash}ctx^+(B_n)$. The hypothesis is initialized using the current noise state for the first beat, $hyp\_noise^l_{n}\!=\!lead\_noise^l_{n-1}$, and the value of the previous beat for each new one, $hyp\_noise^l_{n+i}\!=\!hyp\_noise^l_{n+i-1}$. This value can be further modified for a beat $B_{n+i}$ under three circumstances:

\begin{itemize}
\item If $B_{n+i}$ is considered noisy in lead $l$ by the beat-based noise detection, $beat\_noise^l_{n+i}\!\!=\!true$, then the hypothesis is set to $hyp\_noise^l_{n+i}\!\!=\!true$.
\item If a new cluster is created for $B_{n+i}$, we define $L_{n+i}^{noise}$ as the set of leads with $\bar{\mathcal{S}}(\boldsymbol{q}_{n+i}^{l},\boldsymbol{q}_{n+i-1}^{win,l})\!\le\!\gamma$, which are responsible for its creation and candidates to be set as noisy. For these leads, we check the assignment under noisy interval conditions, $\widebar{\mathcal{P\!S}}(\boldsymbol{q}^l_{n+i},\boldsymbol{q}_{n+i-1}^{win,l})\!\!>\!\!\gamma$. If the assignment condition is now fulfilled in all leads, the signal still resemble a previous QRS morphology and we set $hyp\_noise_{n+i}^l\!\!=\!true, \forall l\!\in\!L_{n+i}^{noise}$.
\item If $B_{n+i}$ is the last of $\kappa$th consecutive noise-free beat, we set $hyp\_noise_{n+i-j}^l\!=\!false, \forall j \in [0,\kappa\!\!-\!\!1]$. 
\end{itemize}
 
Once the $\tau$ beats have been processed, the detection of noisy leads is addressed. Let $C^{ctx-new}_{n+\tau-1}$ denote the set of new clusters created in $\tau\text{\textendash}ctx^+(B_n)$. If $|C_{n+\tau-1}^{ctx-new}|>\tau/3$ we consider that an abnormal cluster proliferation exists since the appearance of more than 5 new QRS morphologies within an interval of 15 beats is an extremely rare condition. Hence any lead $l$ that is a common cause of creation, that is, $l\in L^{noise}_{n+i}$, for all $i$ such that $B_{n+i} \in C_{n+i}^c \wedge C_{n+i}^{c}\!\in\!C^{ctx-new}_{n+\tau-1}$, is discarded as noisy, setting $hyp\_noise^l_{n+i}\!=\!true, \forall i \in [0,\tau\!\!-\!\!1]$.

Finally, the hypothesis is reviewed for every beat $B_{n+i}\!\in\!\tau\text{\textendash}ctx^+(B_n)$ which caused the creation of a new cluster so as to decide about its noisy condition. If $hyp\_noise^l_{n+i}\!\!=\!\!true,$ $\forall l\!\in\!L_{n+i}^{noise}$ then we set $beat\_noise_{n+i}^l\!\!=\!\!true$, the created cluster is deleted and its beats assigned to its closest cluster. Afterwards, $lead\_noise_{n+i}^l$ is updated for the whole $\tau\text{\textendash}ctx^+(B_n)$ using the $beat\_noise_{n+i}^l$ data to set the begining and ending of noise intervals for each lead. The resulting $lead\_noise_{n+\tau-1}^l$ is the new noise state used in the analysis of the subsequent beats.

\section{Rhythm analysis}\label{Rhythm}

The absence of an analysis of the P wave leads to the inability to discriminate premature, normal or ectopic beats which share a common QRS morphology. All previous clustering proposals include rhythm information within their beat characterization and claim its separation capabilities for this kind of arrythmias. In order to make our results comparable, we include a rhythm processing stage that allows us to separate those beats into different groups.

\subsection{$R\!R$-Interval characterization}

\begin{figure}[!t]
\centering
\includegraphics[width=3.49in]{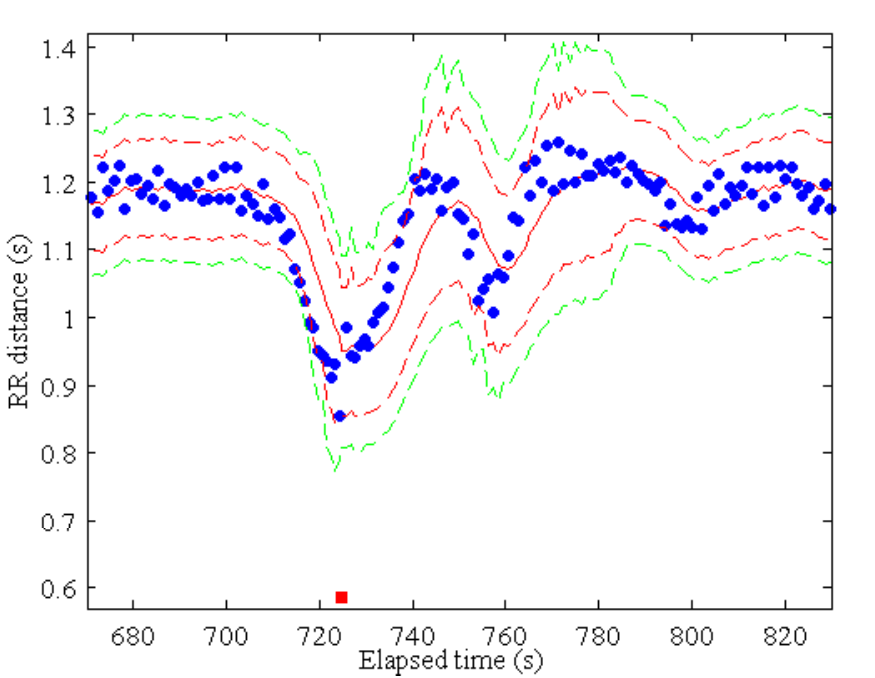}\caption{An excerpt of the temporal evolution of $R\!R$ distance for record 117 of MIT-BIH database is shown. Solid points and squares represent normal and premature beats respectively. The solid line represents $\widehat{N\!N}_n$ and the dotted lines enclose the $\pm 2\widehat \sigma_n$ and $\pm 3\widehat \sigma_n$ intervals.}
\label{fig_RR}
\end{figure}

The beat to beat interval ($R\!R$) between normal beats, commonly known as $N\!N$ interval, is the result of a non-stationary stochastic process regulated by the sympathetic and parasympathetic nervous systems. This implies that the $R\!R$ value for beat $B_n$, denoted by $R\!R_n$, should be put in context using the rhythm of the surrounding beats to analyze its normality. To this end, we model the $N\!N$ series as a stochastic process with marginal distribution $\mathcal{N}(\widebar{N\!N}_n,\sigma_n^2)$ at the $n$th beat. The mean is estimated as $\widehat{N\!N}_{n}\!=\!\theta R\!R_{n-1}\!+\!(1\!-\!\theta)\widehat{N\!N}_{n-1}$, with $\theta\!=\!0.2$ for $R\!R_{n-1}$ values labeled as normal (as explained in next section). Otherwise, $\widehat{N\!N}_{n}\!=\!\widehat{N\!N}_{n-1}$. The standard deviation is estimated as  $\widehat{\sigma}_n^2\!=\!\sum_{i} N\!N_i\!-\!\widehat{N\!N}_{i}$  using the last $\tau$ beats with normal rhythm. Fig. \ref{fig_RR} shows the evolution of the $R\!R$ in an excerpt with a premature beat from the record 117 of the MIT-BIH database to illustrate this point.

The first complete context, $\tau\text{\textendash}ctx^-(B_{\tau+1})$,  is used to initialize $\widehat{N\!N}_{n}$ and $\widehat \sigma_n$. Let $T\!C$ denote the set of the first $\tau$ values of $R\!R$ and let $K_i$ denote any subset of $R\!R$ values from consecutive beats. A value $R\!R_j\!\in\!T\!C$ is said to be normal if $ \exists K_i\!$ such that $\!R\!R_j\!\in\!K_i \wedge \sigma_{K_i}\!<\!0.1$, where $\sigma_{K_i}$ is the normalized standard deviation of the $K_i$ set. Let $T\!C_N\!=\!\{R\!R_j\!\mid\!R\!R_j\text{ is normal}\}$. If $T\!C_N\!=\!\emptyset$, then $T\!C_N\!=\!\{R\!R_j\!\mid\!R\!R_j\!\in\![\widebar {T\!C}\! -\! 2\sigma_{TC}, \widebar {T\!C}\!+\! 2\sigma_{TC}]\}$, where $\widebar {T\!C}$ and $\sigma_{TC}$ are the mean and standard deviation of $TC$. If $T\!C_N\!=\!\emptyset$ then the context in $T\!C$ is repeatedly moved forward one beat until $T\!C_N\!\ne\!\emptyset$ for a context $\tau\text{\textendash}ctx^-(B_{\tau+1+i})$. Finally we set $\widehat{N\!N}_n\!=\!\widebar{T\!C}_N$ and $\widehat \sigma_n\!=\!\sigma_{T\!C_N},$ $\forall n\!\in\![1,\tau\!+1+\!i]$.

\subsection{Rhythm labeling}

The aim of rhythm processing is the discrimination of those abnormal $R\!R$ values associated with arrhythmic beats from the normal ones. To this end, the rhythm of the $B_n$ beat is characterized through a vector $\boldsymbol{rr}_n\!=\!(R\!R_n,R\!R_{n}^-,R\!R_n^+,\widehat{N\!N}_{n},\widehat \sigma_{n})$ where $R\!R_{n}^-$ and $R\!R_n^+$ are the $R\!R$ values for the previous and next beat, respectively. Then, the model described in the previous section is used to establish a range of validity for the $R\!R_n$ value which allows us to detect any alteration in the normal rhythm. 

We use seven rhythm labels to discern between four beat rhythm types: normal, with ($C$) or without ($N$,$N^-$,$N^+$) compensatory pause, premature ($P$), group of prematures ($G\!P$) and delayed ($D$). The explicit domain knowledge contained in Table \ref{ClassifierRules} models, for each rhythm type, the relation of an $R\!R$ value with the normal rhythm from its temporal context. It also reflects the dependence of the rhythm type for an $R\!R$ value on the rhythm type of its adjacent beats. This model allows us to assign a rhythm label to the beat $B_n$ based on the $\boldsymbol{rr}_n$ values and the rhythm label of the previous beat. 

\begin{table}[!t]
\renewcommand{\arraystretch}{1.2}
\renewcommand{\tabcolsep}{2pt}
\caption{Rules for rhythm label selection. $\Delta
R\!R_n$ stands for $R\!R_n\!-\!\widehat{N\!N}_{\!n}$.
The superscripts refer the conditions listed below and used when multiple candidate labels exist. They impose a minimum distance between normal and premature or delayed beats. The precedence of label candidates is set by order of appearance.}
 \label{ClassifierRules} 
 \centering 
  \begin{tabular}{l|c|c|c|c|c|c|c|} 
&\multicolumn{7}{>{\centering}c<{\centering}|}{Rhythm label for the previous beat} \\
\cline{2-8}
$\Delta R\!R_n $	& $\mathbf{GP}$ & $\mathbf{P}$ &$\mathbf{N_-}$ &
$\mathbf{N}$ & $\mathbf{N_+}$ & $\mathbf{C}$ & $\mathbf{D}$ \\\hline

\multirow{2}{*}{$\in (3\widehat{\sigma}_n,\infty] $} 
                & $D^6$	        & $D^{2,6}$    & $D$           & $D^{1|10}$   & $D^{2|11}$     & $D$          & $D$       \\
		& $C$	        & $C$          &               & $N_+$        & $N_+$          &              &           \\\hline

\multirow{2}{*}{$\in (2\widehat{\sigma}_n,3\widehat{\sigma}_n]$}
		& $C$	        & $C$          & $N_+$	       & $N_+$        & $N_+$          & $D^{5,6}$    & $D^{6|9}$ \\
                &               &              &               &              &                & $N_+$        & $N_+$     \\\hline
$\in [-2\widehat{\sigma}_n,2\widehat{\sigma}_n]$  
                & $N$           & $N$          & $N$           & $N$          & $N$            & $N$          & $N$       \\\hline

\multirow{3}{*}{$\in [-3\widehat{\sigma}_n,-2\widehat{\sigma}_n)$}
                & $GP^8$        & $N_-^1$      & $N_-$	       & $P^{3,4,6}$ 
& $N_-$	       & $P^{6}$      & $P^4$ \\
                & $N_-$         & $GP$         &               & $GP^{3,8}$   & 
              & $GP^{8}$     & $N_-$          \\
                &               &              &               & $N_-$        &                & $N_-$        &           \\\hline

\multirow{3}{*}{$\in [-\infty,-3\widehat{\sigma}_n) $}
		& $GP$          & $GP$         & $P^{3,4,7}$   & $P^{3,4,7}$  & $P^{4}$        & $P^{4,7}$    & $P^4$     \\
                &               &              & $GP^{3,12}$   & $GP^{3}$     & $GP$           & $GP$         & $GP$      \\
                &               &              & $N_-$         & $N_-$        &                &              &           \\\hline
  \end{tabular}
\subfloat{
  \begin{tabular}{llcll} 
1. & $R\!R_n > R\!R_n^- + 4 \widehat {\sigma}_n$ && 7. & $R\!R_n^+ > \widehat{N\!N}_n - 3 \widehat {\sigma}_n$\\  
2. & $R\!R_n > R\!R_n^- + 3 \widehat {\sigma}_n$ && 8. & $R\!R_n^+ < \widehat{N\!N}_n - 3 \widehat {\sigma}_n$\\ 
3. & $R\!R_n < R\!R_n^- - 3 \widehat {\sigma}_n$ && 9. & $R\!R_n^+ > \widehat{N\!N}_n - 2 \widehat {\sigma}_n$\\ 
4. & $R\!R_n < R\!R_n^+ - 3 \widehat {\sigma}_n$ && 10. & $R\!R_n^+ > R\!R_n^- + 4 \widehat {\sigma}_n$\\
5. & $R\!R_n^- > \widehat{N\!N}_n + 3 \widehat {\sigma}_n$ && 11. & $R\!R_n^+ > R\!R_n^- + 3 \widehat {\sigma}_n$\\
6. & $R\!R_n^+ > \widehat{N\!N}_n + 3 \widehat {\sigma}_n$ && 12. & $R\!R_n^+ < R\!R_n^- - 3 \widehat {\sigma}_n$\\ 
  \end{tabular}}
\centering 
\end{table} 

\section{Results}\label{Results}

We have applied our clustering method to all the records of the MIT-BIH database. The parameters and threshold values of this method have been neither trained nor adjusted to fit this database. These values have been justified by physiological reasons, or by the expertise or intuition of experienced cardiologists. The method shows low sensitivity to small changes in parameter values; the results either improve or worsen slightly. Although better results could be obtained by a fine tunning of these parameters for each specific database, this is not the aim of this work but proving the validity of the present approach for continuous ECG monitoring. 

For each record, a set of clusters is obtained reflecting the QRS morphologies present in them. Afterwards, the rhythm labels are used to split each cluster into groups of beats with the same rhythm type. In order to compare our results with the proposal in \cite{Lagerholm2000} under equivalent conditions, we adopted a fixed number of clusters (25) as the maximum number of groups. Thus, if that limit is exceeded for a record, a merging process is applied to obtain a reduced set of groups with the most prevalent rhythm and morphologies. If necessary, we keep merging the groups with the lowest number of assigned beats until the maximum is reached. Table \ref{clusters_per_record} shows the results before and after the merging process. 

Each group is labeled with the majority class label of the beats assigned to this group from the database. An assigned beat is considered as correctly grouped if the class label in the database match the label of the group. A confusion matrix is obtained for each record comparing both labels for every beat. These matrices are summed to obtain the global confusion matrix for the whole validation set shown in Table \ref{MIT-BIH_confusion_matrix}. Table \ref{MIT-BIH_confusion_matrix_AAMI} shows the results using the AAMI class labels obtained from MIT-BIH labels as described in \cite{EC57-2008}. Table \ref{AAMI_confusion_matrix} shows the results on AHA database.

\begin{table}[!t] 
\renewcommand{\tabcolsep}{4pt}
\caption{Number of clusters per record in MIT-BIH DB. N column indicates the number of clusters created using QRS morphology; N$_{R\!R}$, the number of groups after using rhythm labels; and N$_{R\!R}^c$, the number of groups after the merging process.} 
 \label{clusters_per_record} 
 \centering 
  \begin{tabular}{c|c|c|c||c|c|c|c} 
 Rec. & N &  N$_{R\!R}$ & N$_{R\!R}^c$ & Rec. & N &  N$_{R\!R}$ & N$_{R\!R}^c$ \\\hline

100 & 4 & 7 & 7 & 
201 & 15 & 32 & 10 \\ 
101 & 4 & 6 & 6 &
202 & 9 & 18 & 18 \\ 
102 & 10 & 13 & 13 & 
203 & 33 & 87 & 25 \\
103 & 10 & 12 & 12 &
205 & 14 & 20 & 20 \\
104 & 16 & 25 & 25 & 
207 & 61 & 96 & 25 \\
105 & 10 & 16 & 16 & 
208 & 28 & 63 & 22 \\
106 & 27 & 49 & 18 & 
209 & 10 & 26 & 9 \\ 
107 & 11 & 21 & 21 &
210 & 27 & 65 & 13 \\
108 & 22 & 35 & 7 &
212 & 5 & 8 & 8 \\
109 & 13 & 18 & 18 &
213 & 17 & 29 & 11\\
111 & 8 & 10 & 10 & 
214 & 21 & 36 & 11\\
112 & 4 & 7 & 7 & 
215 & 16 & 32 & 8\\
113 & 5 & 9 & 9 & 
217 & 28 & 50 & 19\\
114 & 8 & 13 & 13 &
219 & 14 & 24 & 24\\
115 & 11 & 11 & 11 & 
220 & 2 & 5 & 5\\
116 & 10 & 15 & 15 & 
221 & 14 & 24 & 24\\
117 & 4 & 5 & 5 &
222 & 8 & 19 & 19\\
118 & 3 & 8 & 8 & 
223 & 23 & 43 & 17\\
119 & 6 & 13 & 13 &
228 & 14 & 25 & 25\\
121 & 5 & 7 & 7 & 
230 & 3 & 5 & 5\\
122 & 1 & 1 & 1 &
231 & 5 & 6 & 6\\
123 & 3 & 5 & 5 &
232 & 4 & 9 & 9\\
124 & 14 & 18 & 18 &
233 & 24 & 48 & 15\\
200 & 20 & 46 & 16 &
234 & 5 & 7 & 7\\\hline
\end{tabular} 
 \centering 
\end{table} 

\begin{table*}[!t]
\renewcommand{\arraystretch}{1.3}
\renewcommand{\tabcolsep}{4pt}
\caption{Confusion matrix resulting from clustering MIT-BIH Arrhythmia database. The first row corresponds to the annotation labels of the database, and the first column, to the dominant annotation label in the clusters. $\boldsymbol{Se}$, $\boldsymbol{+\!P}$ and $\boldsymbol{F\!P\!R}$ denote the Sensitivity, Positive Predictivity and False Positive Rate for each beat class respectively.}
 \label{MIT-BIH_confusion_matrix} 
 \centering 
  \begin{tabular}{c|c|c|c|c|c|c|c|c|c|c|c|c|c|c|c|c} 

  & N & L & R & a & V & F & J & A & S & E & j & / & e & f & Q & !\\\hline 
\textbf{N}
           & 74618	& 0 	& 3 	& 15 	& 167 	& 108 	& 11 	& 196 	& 2 	& 1 	& 60 	& 1 	& 15 	& 38 	& 9 	& 0\\\hline 
\textbf{L} & 0 	   	& 8059 	& 0 	& 0 	& 1 	& 2 	& 0 	& 0 	& 0 	& 0 	& 0 	& 0 	& 0 	& 0 	& 2 	& 0\\\hline 
\textbf{R} & 32		& 0 	& 7245 	& 0 	& 1 	& 3 	& 29	& 8 	& 0 	& 6 	& 6 	& 0 	& 0 	& 0 	& 0 	& 4\\\hline 
\textbf{a} & 7 		& 0 	& 0 	& 120	& 1 	& 0 	& 1 	& 2 	& 0 	& 0 	& 5 	& 0 	& 0 	& 0 	& 0 	& 0\\\hline 
\textbf{V} & 131 	& 1 	& 0 	& 11 	& 6848	& 83 	& 0 	& 20 	& 0 	& 0 	& 0 	& 0 	& 0 	& 1 	& 4 	& 6\\\hline 
\textbf{F} & 26 		& 0 	& 0 	& 2 	& 56 	& 606 	& 0 	& 2 	& 0 	& 0 	& 0 	& 0 	& 0 	& 0 	& 0 	& 0\\\hline 
\textbf{J} & 1 		& 0 	& 0 	& 0 	& 0 	& 0 	& 42	& 0 	& 0 	& 0 	& 0 	& 0 	& 0 	& 0 	& 0 	& 0\\\hline 
\textbf{A} & 44 		& 4 	& 3 	& 2 	& 25	& 0 	& 0 	& 2317 	& 0 	& 0 	& 0 	& 0 	& 1 	& 0 	& 0 	& 59\\\hline 
\textbf{S} & 0 		& 0 	& 0 	& 0 	& 0 	& 0 	& 0 	& 0 	& 0 	& 0 	& 0 	& 0 	& 0 	& 0 	& 0 	& 0\\\hline 
\textbf{E} & 0 		& 0 	& 2 	& 0 	& 10 	& 0 	& 0 	& 0 	& 0 	& 93 	& 0 	& 0 	& 0 	& 0 	& 0 	& 0\\\hline 
\textbf{j} & 145 	& 0 	& 0 	& 0 	& 0 	& 0 	& 0 	& 0 	& 0 	& 0 	& 158 	& 0 	& 0 	& 0 	& 0 	& 0\\\hline 
\textbf{/} & 9 		& 0 	& 0 	& 0 	& 1 	& 0 	& 0 	& 0 	& 0 	& 0 	& 0 	& 7010 	& 0 	& 127 	& 0 	& 0\\\hline 
\textbf{e} & 0 		& 0 	& 0 	& 0 	& 0 	& 0 	& 0 	& 0 	& 0 	& 0 	& 0 	& 0 	& 0 	& 0 	& 0 	& 0\\\hline 
\textbf{f} & 9 		& 0 	& 0 	& 0 	& 5 	& 0 	& 0 	& 0 	& 0 	& 0 	& 0 	& 13 	& 0 	& 816 	& 15	& 0\\\hline 
\textbf{Q} & 0 		& 0 	& 0 	& 0 	& 0 	& 0 	& 0 	& 0 	& 0 	& 0 	& 0 	& 0 	& 0 	& 0 	& 3 	& 0\\\hline 
\textbf{!} & 0 		& 8 	& 0 	& 0 	& 12	& 0 	& 0 	& 0 	& 0 	& 6 	& 0 	& 0 	& 0 	& 0 	& 0 	& 403\\\hline 
\textbf{Se}& 99.46 & 99.84 & 99.89 & 80.00 & 96.09 & 75.56 & 50.60 & 91.04 & 0.00 & 87.74 & 69.00 & 99.80 & 0.00 & 83.10 & 9.09 & 85.38\\
\textbf{+P}& 99.17 & 99.94 & 98.79 & 88.24 & 96.38 & 87.57 & 97.67 & 94.38 & - & 88.57 & 52.15 & 98.08 & - & 95.10 & 100.00 & 93.94\\
\textbf{FPR}& 1.79 & 0.00 & 0.09 & 0.01 & 0.25 & 0.08 & 0.00 & 0.13 & 0.00 & 0.01 & 0.13 & 0.13 & 0.00 & 0.04 & 0.00 & 0.02\\
   \end{tabular} 
 \centering 
\end{table*} 

\begin{table}[t] 
\renewcommand{\arraystretch}{1.3}
\caption{Confusion matrix for MIT-BIH DB using AAMI class labels.} 
 \label{MIT-BIH_confusion_matrix_AAMI} 
 \centering 
  \begin{tabular}{c|c|c|c|c|c} 
& N & S & V & F & Q\\ 
   \hline 
\textbf{ N }& 89957&    342&    176&      113&     50 \\ 
   \hline 
\textbf{ S }&    204&  2648&    26&      0&      0 \\ 
   \hline 
\textbf{ V }&     134&     31&    6951&     83&     5 \\ 
   \hline 
\textbf{ F }&      26&      4&      56&      606&       0 \\ 
   \hline 
\textbf{ Q }&      18&     0&      6&      0&      7984 \\ 
   \hline 
   \hline 
\textbf{ Se } &  99.58&  87.54&  96.34&  75.56&  99.32\\ 
\textbf{ +P } &  99.25&  92.01&  96.49&  87.57&  99.70\\ 
\textbf{ FPR }&   3.57&   0.22&   0.25&   0.08&   0.02\\ 

  \end{tabular} 
 \centering 
\end{table} 

\begin{table}[!t] 
\renewcommand{\arraystretch}{1.3}
\caption{Confusion matrix for AHA ECG Database. Classes from the MIT-BIH label set without assigned beats are omitted.} 
 \label{AAMI_confusion_matrix} 
 \centering 
  \begin{tabular}{c|c|c|c|c|c|c} 
& N & V & F & E & / & Q \\ 
   \hline 
\textbf{ N }& 318739&    472&    138&      7&     16&    150 \\ 
   \hline 
\textbf{ V }&    229&  32090&    217&      0&      9&     66 \\ 
   \hline 
\textbf{ F }&     33&     115&    885&      0&     16&      1 \\ 
   \hline 
\textbf{ E }&      0&      0&      0&      5&      0&      0 \\ 
   \hline 
\textbf{ / }&      5&      23&     26&      0&   3128&      1 \\ 
   \hline 
\textbf{ Q }&      11&     32&      0&      0&      0&    354 \\ 
   \hline 
   \hline 
\textbf{ Se } &  99.91&  98.04&  69.91&  41.67&  98.71& 61.89\\ 
\textbf{ +P } &  99.75&  98.40&  84.29& 100.00&  98.27& 89.17\\ 
\textbf{ FPR }&   2.07&   0.16&   0.04&   0.00&   0.02&  0.01\\ 
  \end{tabular} 
  
 \centering 
\end{table}

\subsection{Real-time considerations}

The proposed method processes an ECG recording with a bounded time delay comprising the intrinsic latency and computation time for the different stages. Only the baseline filtering and rhythm labeling stages present an intrinsic latency due to non-causality: $400$ms and one beat, respectively. Since they both are executed concurrently, their contribution to the delay is given by the maximum of both. The computation time can be evaluated through the time complexity which is constant in all stages but two: QRS alignment and template matching. In both cases, the time complexity is constant for the best case, corresponding to beats assigned to a cluster in the context; this happens in 95.35\% of the total number of beats in MIT-BIH database. The time complexity is linear ($\mathcal{O}(|C_n|)$) for the worst case, which corresponds to beats assigned to a new or an out-of-context cluster, representing the remaining 4.65\% of the total number of beats. Given the high degree of parallelism in both stages and the computational cost of processing a single cluster, it can be guaranteed that a beat is clustered before the next one arrives even with a set of hundreds of clusters. In order to support this claim, the MIT-BIH Arrhythmia database was processed with a non-optimized, non-parallelized MATLAB implementation of the method using a single core of an Intel Q9550 CPU. The following computation times for a single beat were obtained: QRS characterization (maximum, mean): 4ms, 3ms; cluster set selection: 323ms, 16ms; cluster set updating: 110ms, 5ms; noise analysis: 9ms, $<$1ms; and rhythm-based labelling: 10ms, 5ms. Globally, the whole method summed a maximum of 358ms and mean of 32ms. Additionally, reducing groups for validation required a maximum of 190ms.

\subsection{Clustering performance measures}

Purity is usually used to measure the goodness of a clustering method. Nevertheless, in a multiclass problem like this one, after characterizing the clusters, the values of sensitivity (Se), positive predictivity (+P), and false positive rate (FPR) should also be provided for each class, in order to obtain a multidimensional measure of the quality of the results from the perspective of each class involved. A global purity of 98.56\% is obtained for MIT-BIH Arrhythmia database (98.84\% with AAMI class labels) and a 99.56\% for the AHA ECG database. The other values are shown in the last row of Tables \ref{MIT-BIH_confusion_matrix}\textendash \ref{AAMI_confusion_matrix}.

\section{Discussion and Conclusion}\label{Discussion}

We propose a new clustering method to dynamically separate QRS morphologies as they appear in a multichannel ECG signal, representing them with a dynamic number of clusters. This objective has not been previously addressed in the literature and only some partial solutions can be found, all of them restricted to offline processing \cite{Lagerholm2000,CuestaFrau2003,CuestaFrau2007} and some of them even limited to single channel signals and to a fixed subset of beat classes \cite{CuestaFrau2003,CuestaFrau2007}.

The performance of the QRS clustering technique without using rhythm data shows a global purity of 97.15\% and 99.43\% for MIT-BIH and AHA databases, respectively. This confirms the validity of our approach, since no method, as far as we know, neither offline nor online, achieve this performance without using $R\!R$ derived information. As expected, the main source of error is the group of supraventricular classes A, N, J, j, and e, that can only be separated using P wave and rhythm information. This is the reason for the difference between both results since AHA database does not contain this type of beats.

The validation results after using rhythm labels show a high sensitivity and positive predictivity for almost all classes while the purity increases. We observe in accordance with \cite{Lagerholm2000}, that the largest number of errors in Table \ref{MIT-BIH_confusion_matrix} are caused by beats with similar morphology. Fusion (F) of ventricular (V) and normal (N) beats are included with N or V clusters and viceversa. The same occurs with N, paced (/) and fusion of N and / beats (f). Finally, beats with supraventricular or nodal activation points (A, N, J, j, and e) with similar QRS are wrongly clustered when the rhythm information does not result determinant for their discrimination. This kind of errors represent 66\% of the total.

Besides the clustering performance, the analysis of the number of clusters (N) generated for each record shows that it remains reduced: 34 records (71\%) have 15 or less clusters; 12 records (25\%) have between 16 and 30 clusters, both included and only two records (4\%) has more than 30 clusters. The high number of clusters for record 207 is caused by the presence of an episode of ventricular flutter with QRS complexes replaced by irregular waves. The clustering results would be improved if an specific detection method were used for these kind of arrhythmias with absent QRS complex.

The $N_{R\!R}$ column of Table \ref{clusters_per_record} shows a general increase in the number of groups in all records with a mean of 24 groups per record. This increment reflects the presence of different rhythm labels in the beats assigned to the clusters although they do not always belong to different beat classes. Records with irregular rhythm like those with auricular fibrillation, or sudden rhythm change will render the $R\!R$ information useless to discriminate premature or delayed beats. In such cases, the $R\!R$ does not provide information about the beat activation point and the discrimination cannot be performed without an analysis of the P wave.

Since no other method for online clustering has been reported, we compare our clustering performance with existing offline proposals. Only the work of Lagerholm et al.\cite{Lagerholm2000} provides comparable results since others \cite{CuestaFrau2003,CuestaFrau2007} are designed to deal with a concrete subset of beat types and perform the
evaluation over a single channel (usually the one with lower noise). Compared to \cite{Lagerholm2000} our method provides a slightly better purity (98.56\% vs
98.49\%). The sensitivity is improved in our work for 11 out of the 16 beat classes and slightly worsened for 3. Special mention should be made for classes a, A, R, j and E where the improvement is remarkable. Let us remember that \cite{Lagerholm2000} rely on a SOM with 25 clusters to represent the different beat classes. This approach has two main drawbacks, first the clusters get saturated with dominant morphologies present in the learning stage, while rare morphologies are ignored as well as new morphologies that appear afterwards. Second, the generated clusters are redundant in those records with a low number of morphologies. In contrast, our method dynamically adapts the number of clusters to the number of morphologies detected.

The results of our proposal confirms the relevance of the temporal context for beat clustering. It allows us to switch from an offline to an online analysis achieving the same or even better results, and to address the temporal evolution of a beat morphology that otherwise would be projected into multiple clusters. The results of the experiments on ECG standard databases also show the adequacy of the present method for real-time ECG monitoring.

Our proposal provides the cardiologists with the information about the morphological diversity within a desired time frame and its temporal evolution. This information allows them to promptly detect the different conduction patterns and evaluate its relevance. It also can be useful for arrhythmia detection and classification which can be later addressed either automatically by classification algorithms or manually by the cardiologists. 

In conclusion, we have presented an adaptive, multichannel context-based method for clustering beat morphologies in real-time that has been validated over the whole MIT-BIH Arrhythmia and AHA ECG databases with performance results that outperform its offline counterparts in the field.

\ifCLASSOPTIONcaptionsoff
  \newpage
\fi

\bibliographystyle{IEEEtran}

\begin{thebibliography}{23}
\providecommand{\url}[1]{#1}
\csname url@samestyle\endcsname
\providecommand{\newblock}{\relax}
\providecommand{\bibinfo}[2]{#2}
\providecommand{\BIBentrySTDinterwordspacing}{\spaceskip=0pt\relax}
\providecommand{\BIBentryALTinterwordstretchfactor}{4}
\providecommand{\BIBentryALTinterwordspacing}{\spaceskip=\fontdimen2\font plus
\BIBentryALTinterwordstretchfactor\fontdimen3\font minus
  \fontdimen4\font\relax}
\providecommand{\BIBforeignlanguage}[2]{{\expandafter\ifx\csname l@#1\endcsname\relax
\typeout{** WARNING: IEEEtran.bst: No hyphenation pattern has been}\typeout{** loaded for the language `#1'. Using the pattern for}\typeout{** the default language instead.}\else
\language=\csname l@#1\endcsname
\fi
#2}}
\providecommand{\BIBdecl}{\relax}
\BIBdecl

\bibitem{Sutton2013}
R.~Sutton, ``Remote monitoring as a key innovation in the management of cardiac
  patients including those with implantable electronic devices,''
  \emph{Europace}, vol.~15, no. suppl 1, pp. i3--i5, 2013.

\bibitem{DeChazal2004}
P.~DeChazal, M.~O'Dwyer, and R.~B. Reilly, ``Automatic classification of
  heartbeats using {ECG} morphology and heartbeat interval features,''
  \emph{IEEE Trans. Biomed. Eng.}, vol.~51, no.~7, pp. 1196--1206, 2004.

\bibitem{Llamedo2011}
M.~Llamedo and J.~P. Mart{\'i}nez, ``Heartbeat classification using feature
  selection driven by database generalization criteria,'' \emph{IEEE Trans.
  Biomed. Eng.}, vol.~58, no.~3, pp. 616--625, 2011.

\bibitem{deLannoy2012}
G.~de~Lannoy, D.~Fran\c{c}ois, J.~Delbeke, and M.~Verleysen, ``Weighted
  conditional random fields for supervised interpatient heartbeat
  classification,'' \emph{IEEE Trans. Biomed. Eng.}, vol.~59, no.~1, pp.
  241--247, 2012.

\bibitem{Lagerholm2000}
M.~Lagerholm, C.~Peterson, G.~Braccini, L.~Edenbrandt, and L.~Sornmo,
  ``Clustering {ECG} complexes using {H}ermite functions and self-organizing
  maps,'' \emph{IEEE Trans. Biomed. Eng.}, vol.~47, no.~7, pp. 838--848, 2000.

\bibitem{CuestaFrau2003}
D.~Cuesta-Frau, J.~C. P{\'e}rez-Cortes, and G.~A. Garc{\'i}a, ``Clustering of
  electrocardiograph signals in computer-aided holter analysis,'' \emph{Comput.
  Meth. Prog. Bio.}, vol.~72, no.~3, pp. 179--196, 2003.

\bibitem{CuestaFrau2007}
D.~Cuesta-Frau, M.~O. Biagetti, R.~A. Quinteiro, P.~Mic{\'o}-Tormos, and
  M.~Aboy, ``Unsupervised classification of ventricular extrasystoles using
  bounded clustering algorithms and morphology matching,'' \emph{Med. Biol.
  Eng. Comput.}, vol.~45, no.~3, pp. 229--239, 2007.

\bibitem{RodrguezSotelo2009}
J.~L. Rodr{\'i}guez-Sotelo, D.~Cuesta-Frau, and G.~Castellanos-Dom{\'i}nguez,
  ``Unsupervised classification of atrial heartbeats using a prematurity index
  and wave morphology features,'' \emph{Med. Biol. Eng. Comput.}, vol.~47,
  no.~7, pp. 731--741, 2009.

\bibitem{Krasteva2007}
V.~Krasteva and I.~Jekova, ``\BIBforeignlanguage{ENG}{{QRS} template matching
  for recognition of ventricular ectopic beats},''
  \emph{\BIBforeignlanguage{ENG}{Ann. Biomed. Eng.}}, Sep 2007.

\bibitem{Christov2006}
I.~Christov, G.~G{\'o}mez-Herrero, V.~Krasteva, I.~Jekova, A.~Gotchev, and
  K.~Egiazarian, ``Comparative study of morphological and time-frequency {ECG}
  descriptors for heartbeat classification,'' \emph{Med. Eng. Phys.}, vol.~28,
  no.~9, pp. 876--887, 2006.

\bibitem{Osowski2001}
S.~Osowski and T.~H. Linh, ``{ECG} beat recognition using fuzzy hybrid neural
  network,'' \emph{IEEE Trans. Biomed. Eng.}, vol.~48, no.~11, pp. 1265--1271,
  2001.

\bibitem{Osowski2004}
S.~Osowski, L.~T. Hoai, and T.~Markiewicz, ``Support vector machine-based
  expert system for reliable heartbeat recognition,'' \emph{IEEE Trans. Biomed.
  Eng.}, vol.~51, no.~4, pp. 582--589, 2004.

\bibitem{Hu1997}
Y.~H. Hu, S.~Palreddy, and W.~J. Tompkins, ``A patient-adaptable {ECG} beat
  classifier using a mixture of experts approach,'' \emph{IEEE Trans. Biomed.
  Eng.}, vol.~44, no.~9, pp. 891--900, 1997.

\bibitem{Dokur2001}
Z.~Dokur and T.~Ölmez, ``Ecg beat classification by a novel hybrid neural
  network,'' \emph{Comput. Meth. Prog. Bio.}, vol.~66, no. 2-3, pp. 167--181,
  2001.

\bibitem{EC57-2008}
\emph{Testing and Reporting Performance Results of Cardiac Rhythm and
  ST-segment Measurement Algorithms}.\hskip 1em plus 0.5em minus 0.4em\relax
  ANSI/AAMI Standard EC57:1998/(R)2008.

\bibitem{Goldberger2000}
A.~L. Goldberger, L.~A.~N. Amaral, L.~Glass, J.~M. Hausdorff, P.~C. Ivanov,
  R.~G. Mark, J.~E. Mietus, G.~B. Moody, C.-K. Peng, and H.~E. Stanley,
  ``{PhysioBank, PhysioToolkit, and PhysioNet}: Components of a new research
  resource for complex physiologic signals,'' \emph{Circulation}, vol. 101,
  no.~23, pp. e215--e220, 2000.

\bibitem{Moody2001}
G.~Moody and R.~Mark, ``The impact of the {MIT-BIH Arrhythmia Database},''
  \emph{IEEE Eng. Med. Biol. Mag.}, vol.~20, no.~3, pp. 45--50, 2001.

\bibitem{Wu2003}
W.-Y. Wu, ``An adaptive method for detecting dominant points,'' \emph{Pattern
  Recogn.}, vol.~36, no.~10, pp. 2231--2237, 2003.

\bibitem{CSE1985}
{The CSE WORKING PARTY}, ``Recommendations for measurement standards in
  quantitative electrocardiography,'' \emph{Eur. Heart. J.}, vol.~6, no.~10,
  pp. 815--825, 1985.

\bibitem{EC13-2007}
\emph{Cardiac Monitors, Heart Rate Meters, and Alarms}.\hskip 1em plus 0.5em
  minus 0.4em\relax ANSI/AAMI Standard EC13:2002/(R)2007.

\bibitem{Kligfield2007}
P.~Kligfield, L.~S. Gettes, J.~J. Bailey, R.~Childers, B.~J. Deal, E.~W.
  Hancock, G.~van Herpen, J.~A. Kors, P.~Macfarlane, D.~M. Mirvis, O.~Pahlm,
  P.~Rautaharju, and G.~S. Wagner, ``Recommendations for the standardization
  and interpretation of the electrocardiogram: Part i.'' \emph{Circulation},
  vol. 115, no.~10, pp. 1306--1324, 2007.

\bibitem{Sakoe1978}
H.~Sakoe and S.~Chiba, ``Dynamic programming algorithm optimization for spoken
  word recognition,'' \emph{IEEE Trans. Acoust., Speech, Signal Process.},
  vol.~26, no.~1, pp. 43--49, 1978.

\bibitem{Keogh2001}
E.~J. Keogh and M.~J. Pazzani, ``Derivative dynamic time warping,'' in
  \emph{1st SIAM Int. Conf. on Data Mining, Chicago, IL, USA}, 2001.

\end{thebibliography}

\begin{IEEEbiographynophoto}{Daniel Castro}
received the B.Sc and M.Sc. degrees in Physics from the University of Santiago de Compostela, Santiago de Compostela, Spain, in 1998 and 1999, respectively. He is currently a researcher and Ph.D. candidate at the Centro Singular de Investigaci\'on en Tecnolox\'ias da Informaci\'on (CiTIUS), University of Santiago de Compostela. His research interests include signal processing and its application to biomedical signals.
\end{IEEEbiographynophoto}
\begin{IEEEbiography}[{\includegraphics[width=1in,height=1.25in,clip,keepaspectratio]{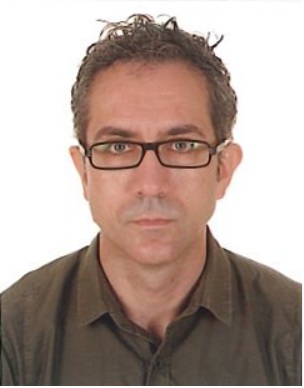}}]{Paulo F\'elix} received the B.Sc and Ph.D. degrees in Physics from the University of Santiago de Compostela, Santiago de Compostela, Spain, in 1993 and 1999, respectively. He is currently a researcher at the Centro Singular de Investigaci\'on en Tecnolox\'ias da Informaci\'on (CiTIUS), University of Santiago de Compostela. His research interests include temporal reasoning, machine learning and signal processing.
\end{IEEEbiography}
\begin{IEEEbiography}[{\includegraphics[width=1in,height=1.25in,clip,keepaspectratio]{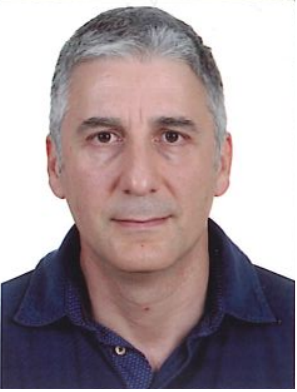}}]{Jes\'us Presedo} received the B.Sc and Ph.D. degrees in Physics from the University of Santiago de Compostela, Santiago de Compostela, Spain, in 1989 and 1994, respectively. He is currently a researcher at the Centro Singular de Investigaci\'on en Tecnolox\'ias da Informaci\'on (CiTIUS), University of Santiago de Compostela. His research interests are biomedical digital signal processing, heart rate variability, non-linear dynamics, soft computing and the development of ubiquitous healthcare systems.
\end{IEEEbiography}

\vfill

\end{document}